\documentclass[pdflatex,sn-mathphys]{sn-jnl}
\usepackage{multicol}
\usepackage{subcaption}
\usepackage[binary-units=true]{siunitx}

\jyear{2022}%

\theoremstyle{thmstyleone}%

\theoremstyle{thmstyletwo}%

\theoremstyle{thmstylethree}%

\begin{document}

\title[FIC: A Framework for
Interaction Robotics]{Fractal Impedance for Passive Controllers: A Framework for
Interaction Robotics}

\author*[1]{\fnm{Keyhan} \sur{Kouhkiloui~Babarahmati}}\email{keyhan.kouhkiloui@ed.ac.uk}
\equalcont{These authors contributed equally to this work.}

\author*[1,2]{\fnm{Carlo} \sur{Tiseo}}\email{c.tiseo@sussex.ac.uk}
\equalcont{These authors contributed equally to this work.}

\author[1]{\fnm{Joshua} \sur{Smith}}\email{s1686489@sms.ed.ac.uk}
\author[3]{\fnm{Hsiu-Chin} \sur{Lin}}\email{hsiu-chin.lin@cs.mcgill.ca}
\author[4]{\fnm{Mustafa} \sur{Suphi~Erden}}\email{m.s.erden@hw.ac.uk}
\author[1]{\fnm{Michael} \sur{Mistry}}\email{michael.mistry@ed.ac.uk}

\affil*[1]{\orgdiv{School of Informatics}, \orgname{University of Edinburgh}, \orgaddress{\street{10 Crichton St}, \city{Edinburgh}, \postcode{EH8 9AB}, \country{UK}}}

\affil[2]{\orgdiv{School of Engineering and Informatics}, \orgname{University of Sussex}, \orgaddress{\street{Chichester 1 Room 002}, \city{Brighton}, \postcode{BN1 9QJ}, \country{UK}}}

\affil[3]{\orgdiv{School of Computer Science/Department of Electrical and Computer Engineering}, \orgname{McGill University}, \orgaddress{\street{3480 Rue University}, \city{Montréal}, \postcode{H3A 2A7}, \state{Quebec}, \country{Canada}}}

\affil[4]{\orgdiv{School of Engineering and Physical Sciences}, \orgname{Heriot-Watt University}, \orgaddress{\city{Edinburgh}, \postcode{EH14 4AS}, \country{UK}}}

\abstract{There is increasing interest in control frameworks capable of moving robots from industrial cages to unstructured environments and coexisting with humans. Despite significant improvement in some specific applications (e.g., medical robotics), there is still the need for a general control framework that improves interaction robustness and motion dynamics. Passive controllers show promising results in this direction; however, they often rely on virtual energy tanks that can guarantee passivity as long as they do not run out of energy. In this paper, a Fractal Attractor is proposed to implement a variable impedance controller that can retain passivity without relying on energy tanks. The controller generates a Fractal Attractor around the desired state using an asymptotic stable potential field, making the controller robust to discretization and numerical integration errors. The results prove that it can accurately track both trajectories and end-effector forces during interaction. Therefore, these properties make the controller ideal for applications requiring robust dynamic interaction at the end-effector.}


\maketitle

\section{Introduction}
\label{sec:Intro}
Robots have been traditionally deployed in highly structured environments with either minimal or heavily controlled human-robot interaction. Their introduction into industries, such as health care, with complex interactive behaviour has revealed the limitations of traditional industrial controllers. The impedance controller \cite{impCtrl, Hogan2014,capelli2022passivity,lachner2021energy,lachner2022shaping,ramuzat2022passive} is a widespread technique enabling robots to interact with uncertain environments. This control technique relies on inverse dynamics modelling to drive the robot to act with a desired mechanical impedance, such as a linear Mass-Spring-Damper system \cite{impCtrl,smith2002synthesis}. Nevertheless, the controller's stability still highly depends on adequate gain tuning, which may be challenging for dynamically intensive tasks. Examples are environments that require adaptive trajectories and/or variable impedance gains \cite{forceImpTrajLearning}, as well as tasks with uncertain end-effector contact against other agents or the environment (e.g., polishing, physical human-robot collaboration, etc.) \cite{varImpCtrlBasedOnHumanStiffEstimation, lin2018projected, HumanLikeAdaptOfForceAndImp,SU2020,Spyrakos-Papastavridis2020,capelli2022passivity,lachner2021energy,lachner2022shaping,ramuzat2022passive}. Such tasks pose various challenges to robots' controllers that currently require an accurate model of contact conditions to ensure system stability.  
\par Variable impedance controllers have been widely explored to address these issues. Ensuring stability with time-variant gains is non-trivial \cite{forceImpTrajLearning,varImpCtrlReinforcementLearApproach,varImpCtrlBasedOnHumanStiffEstimation,TankBasedApproachImpCtrlVarStiff,Spyrakos-Papastavridis2020}; the stability of the system depends not only on the gain profile selected but also on how they are updated. The intrinsic unpredictability of unstructured environments further complicates the process. The variable stiffness controller proposed in \cite{impCtrlWithVarStiffGains} addresses stability but it is not robust to unknown external perturbations. Iterative and adaptive control methods have also been proposed to compensate for the external perturbations by guaranteeing interaction stability \cite{HumanLikeAdaptOfForceAndImp,AVersatileBiomimeticCtrl,zhao2020asymmetrical,dong2019adaptive}. These learning methods are task-specific and do not allow the user any tuning authority on these profiles. Another option to obtain more dexterous motion relies on using force/torque feedback from the end-effector, which is not always viable and is extremely susceptible to vibrations \cite{ForceTrackImpCtrl,S1,S2,S5}. In synthesis, the main issue is that these controllers require accurate models of the external dynamics for stability. Therefore, the robot's stability is highly dependent on environmental modelling, which is difficult to obtain in real-world scenarios. 
\par
Passive controllers have been presented as a viable solution to these problems since their stability is independent of the environmental interaction, in most practical cases \cite{bilateralTeleManWithTimeDelays,portBasedAsymptoticCurveTracking,energyTankBasedWrenchImpCtrl,PowerFlowRegulationAdaptLearnVirtEnergyTank,ramuzat2022passive}. Passive systems are stable because they do not produce energy, but redistribute it at a cost. 
Virtual energy tanks have also been implemented to ensure stability of an active controller by measuring and storing the non-conservative energy via a virtual spring (i.e., integrator) \cite{positionDriftCompensation,bilateralTeleManWithTimeDelays,portBasedAsymptoticCurveTracking,energyTankBasedWrenchImpCtrl,khoramshahi2020dynamical,PowerFlowRegulationAdaptLearnVirtEnergyTank}. Thus, their effectiveness is dependent on the ability of accurately tracking the energy exchanged by the controllers' non-conservative elements (e.g., damping), and their performances are heavily challenged during adaptation to highly variable environments that may consume all the energy accumulated in their tanks \cite{unifiedPassivityBased,passivationOfProjectionBased,stabilityConsiderationsVariImpCtrl}. Energy tank controllers have also been deployed to define energy/power-based safety metrics that allow tuning of the robot impedance, as reported for 1-DoF and multi-DoF platforms \cite{combiningEnergyAndPowerBasedSafety,safetyAndEnergyAwareImpCtrl}. 
Another benefit of passive controllers is their robustness to loss of information over data transmission and system discretisation  \cite{stramigioli2005}.
The main limitations of virtual energy tank controllers are the performance dependency on the residual energy in their tanks and the need for an integrator to account for non-conservative components. The latter becomes relevant in low-bandwidth controllers due to the degraded accuracy of discrete integration. 
\par
\begin{figure}[!tb]
\centering
    \centering
	\includegraphics[width=0.7\columnwidth]{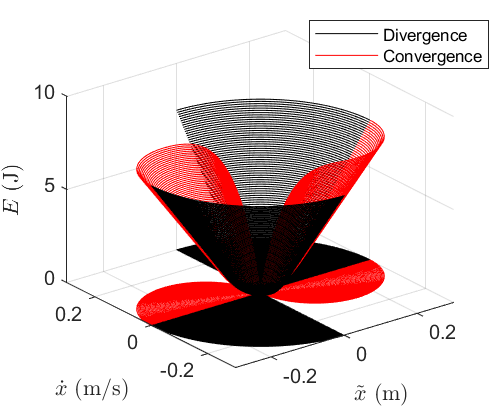}
	\caption{Autonomous trajectories in the phase space of a Fractal Attractor.}
    	\label{fig:1}
\end{figure}

This paper proposes a passive impedance controller where its anisotropic behaviour generates a stable attractor around the desired robot pose, shown in \autoref{fig:1}. The main contributions of the proposed approach are:
\begin{enumerate}
    \item Introducing the Fractal Attractor to design an asymptotically stable impedance controller.
    \item Removing the integrator required to achieve controller passivity in variable stiffness impedance controllers.  
    \item Demonstrating robustness of the proposed controller to low-bandwidth feedback. 
\end{enumerate}
\par
The manuscript is organised as follows: the Method discusses the design of the proposed controller; The Experimental Validation describes the experiments conducted using a Panda 7-DoF Arm (Franka Emika GmbH,DE). The Results provides the results of the experiments. Finally, we analyse the results and draw the conclusions in Discussion and Conclusions, respectively.

\section{Method} \label{sec:Methodology}
The Fractal Impedance Controller (FIC) is a new approach to passive control, which relies upon a state-dependent impedance profile and a Fractal Attractor. The impedance profile determines the robot force and trajectory tracking characteristics, while the Fractal Attractor guarantees smooth autonomous trajectories and stability. In other words, the chosen impedance profile determines the effort the controller opposes to perturbations of the system state. The Fractal Attractor ensures a safe recovery after the perturbation terminates by identifying the harmonic trajectory capable of consuming all the energy accumulated in the Impedance Stiffness ($K_\text{d}$) during divergence.

\subsection{State-dependent Variable Impedance} \label{sec: stateDepVarImpCtrl}

A Cartesian impedance controller drives the end-effector of a robot to generate desired system dynamics \cite{Hogan2014,impCtrl}, such as the following:
\begin{equation} \label{eq:1}
\begin{array}{rl}
{\Lambda}_d\ddot{\tilde{X}} + D_d\dot{\tilde{X}} + K_d\tilde{X} = W_{\text{ext}} + W_{\text{ID}},
\end{array}
\end{equation}
\noindent where ${\Lambda}_d$, $D_d$ and $K_d$ are the desired Cartesian inertia, damping and stiffness matrices, respectively. Note that within the context of this paper, we use upper-case letters to indicate vectors and matrices, while the corresponding lower-case letters (e.g.,\ $k_d$) represent a scalar  component within the vector or matrix.  $W_{\text{ext}}$ is the external wrench applied to the end-effector, and $W_{\text{ID}}$ is the inverse dynamics compensation. $\tilde{X} = X_d - X$ is the error between the desired pose $X_d$ and the current pose $X$, and $\dot{\tilde{X}}$, $\ddot{\tilde{X}}$ are the first and second time derivatives of $\tilde{X}$, respectively.

\begin{figure}
    \centering
	\includegraphics[width=.7\columnwidth]{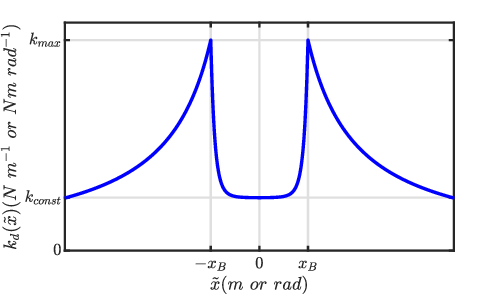}
    \caption{nonlinear stiffness profile used in the fractal impedance controller. Stiffness starts at $k_\text{const}$ near zero and exponentially increases until reaching maximum stiffness $k_\text{max}=w_\text{max}/x_B$. Beyond $x_B$, stiffness decays in order to maintain a maximal force. }
   \label{fig:2}
\end{figure}

Suppose we consider a passive controller without reference velocity ($\dot{\tilde{X}}_d=\mathbf{0}$) and acceleration ($\ddot{\tilde{X}}_d=\mathbf{0}$), which implies that the controller is always pulling the robot toward an equilibrium point. This assumption is essential for system stability in all control architectures that need to be within a region of attraction of a stable point, but it is generally accounted for at the planning stage \cite{Sciavicco}. As a consequence, such a controller would not rely on the planner for stability but may compromise tracking performance. The introduction of a state dependant nonlinear stiffness profile compensates for the loss of the tracking performance by generating a virtual boundary surrounding the desired pose and autonomously adjusting the robot's rigidity to control task accuracy.  The proposed stiffness profile for a single DoF of task-space is as follows: 
\begin{equation} \label{eq:2}
    \begin{array}{l}
            {k_d}(\tilde{x}) = k_{\text{const}} + k_{\text{var}}(\tilde{x}) \\\\
    k_{\text{var}}= \left\{\begin{array}{ll}
	\cfrac{w_\text{max}}{\lvert{\tilde{x}}\rvert} - k_{\text{const}}, \  	& \ \text{if} \ \lvert \tilde{x} \rvert > x_{\text{B}} \\\\
	{e}^{({\beta} {\tilde{x}})^2}, & \  \text{o/w}
	\end{array}\right.\\\\
	{\beta}^2 =\frac{\ln \left( k_{\text{max}}-k_{\text{const}}\right)}{x_{{B}}^2} =\cfrac{\ln \left( \cfrac{w_{\text{max}}}{x_{{B}}}-k_{\text{const}}\right)}{x_{{B}}^2}
	\end{array}
\end{equation}  
\noindent where $k_{d}$ is the $i^{\text{th}}$-element of the block diagonal stiffness matrix  $K_{d}\in \mathbb{R}^{6\times 6}$. $k_\text{const}$ is a fixed constant stiffness. $w_{\text{max}}$ is the maximum exertable task-space force or torque for a single DoF and is determined by the robot's physical properties. $x_{\text{B}}$ is the virtual boundary point where the exerted force saturates. As shown in \autoref{fig:2} the proposed nonlinear stiffness profile starts with a linear region around zero, followed by an exponential growth until the force saturation is reached at $\tilde{x}=x_\text{B}$ and, consequentially, the stiffness starts to decrease with the increase of the displacement in order to maintain a constant maximal force.

In summary, the proposed controller works with the following assumptions: 
\begin{enumerate}
    \item The desired velocity vector ($\dot{X}_\text{d}$) and acceleration ($\ddot{X}_\text{d}$) are equal to $0$, determining that every reference state of the controller is an equilibrium point.
    \item ${\Lambda}_d$ is equal to the task-space inertia of the robot. 
    \item The spring potential energy ($E$) function which is unbounded, explicitly depends only on pose errors (i.e., $E=f(\tilde{X})$).
    \item The controller energy is a uniform continuous function which has a bounded derivative (i.e., a Lipschitz function).
\end{enumerate}

\subsection{Fractal Attractor} \label{sec: onlineAdaptPassiveEnergyTanks}
The Fractal Attractor generates a global region of attraction surrounding the desired state and determines the smooth autonomous trajectories of the controller. It is worth mentioning that Fractal Attractor is the name we give to the stable dynamics generated by the proposed method. The name was chosen because it is a strange attractor. These systems have a fractal structure due to their state dimensionality being a non-integer rational number \cite{strogatz2018nonlinear}, hence the name Fractal Attractor. The proposed system differs from the traditional formulations of strange attractors because it is defined in an algorithmic form rather than parametric nonlinear dynamics equations. This implementation allows changing the system dynamics by selecting a different force profile, resulting in more intuitive programming of the interaction dynamics.

\begin{figure}[!tb]
	\centering
    \begin{subfigure}[b]{.5\columnwidth}
	\includegraphics[width=\columnwidth]{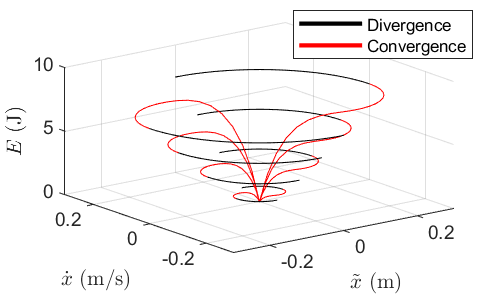}
	\subcaption{\centering\label{fig:3a} Energy of four trajectories of the Fractal Attractor.}
	\end{subfigure}
	\begin{subfigure}[b]{.5\columnwidth}
	\includegraphics[width=\columnwidth]{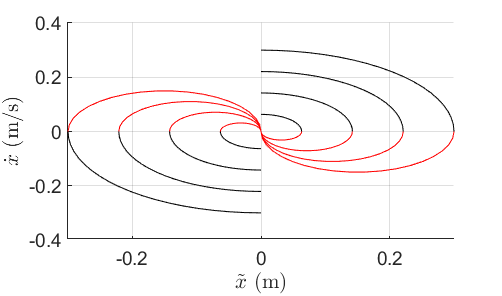}
	\subcaption{\centering\label{fig:3b} Four trajectories of the Fractal Attractor.}
	\end{subfigure}
	\begin{subfigure}[b]{.5\columnwidth}
	\includegraphics[width=\columnwidth]{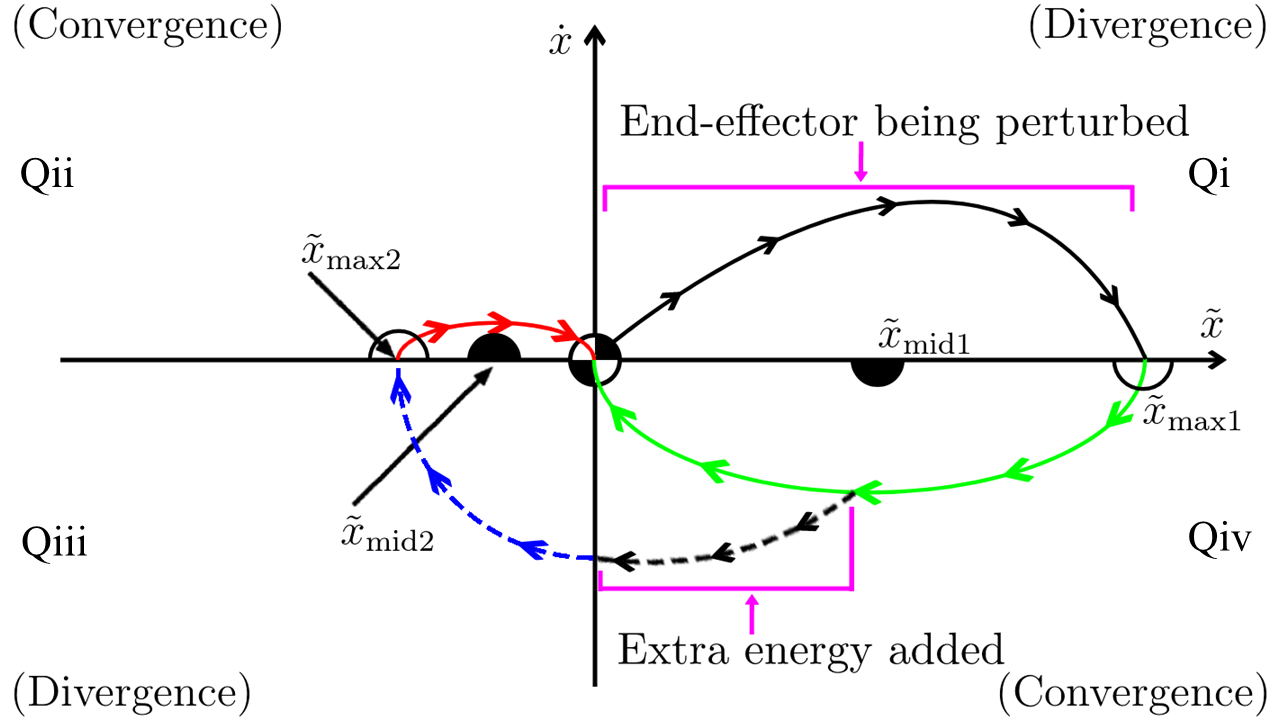}
	\subcaption{\centering\label{fig:3c}: Sample Trajectory}
	\end{subfigure}
	\caption{(a) Energy associated with autonomous trajectories of the Fractal Attractor  (b) The phase portraits show that the attractor topology scales based on the energy accumulated in the controller spring, but they do not change in shape. (c) Starting from the first quadrant of the Cartesian plane $Qi$: the environment perturbs  the end-effector. When the end-effector is released and inverts its motion ($Qiv$), a controller switch is triggered with impedance matched to restore the end-effector to the origin (\textit{green line}). However, if extra energy is added from the environment (\textit{dashed black line}), the trajectory will move into $Qiii$, triggering another controller switch and restoring the original impedance. When the trajectory reaches $Qii$, impedance is again matched to extract the residual energy.}
	\label{fig:3}
\end{figure}

A way to demonstrate that the system is a fractal is to verify that the point-wise dimensionality of its Cantor set is a rational number \cite{strogatz2018nonlinear}.  The point-wise dimensionality is determined numerically by counting the number of states that are progressively encompassed by gradually increasing an isotropic neighbourhood of a chosen point in the system phase-state \cite{strogatz2018nonlinear}. The point-wise dimensionality for the proposed method (\autoref{fig:3}) is $N_{(\tilde{x}=0,~\dot{x}=0)}\propto\left(.5\epsilon^2+\right.$ $\left..5\epsilon^{2-\gamma},~ \gamma \in (0,1]\subset\mathbb{R}\right)$, where $\epsilon$ is the radius of an isotropic neighbour of the considered state. It is smaller than the two-dimensional Cartesian space ($N_{(\tilde{x}=0,~\dot{x}=0)}\propto\epsilon^2$) \cite{strogatz2018nonlinear}.
\begin{proof}
 The proposed attractor has the same dimensionality of the Cartesian Space during divergence (\textit{Qi} and \textit{Qiii} in \autoref{fig:3c}), where it acts as an oscillator around the null state (i.e., $\tilde{x}=0,~\dot{x}=0$), implying that a point for every state enters the neighbourhood every time is increased. In contrast, it has a smaller dimensionality during convergence(\textit{Qii} and \textit{Qiv} in \autoref{fig:3c}), where the attractor trajectory depends by the residual displacement at the at the switching condition ($\dot{x}=0$). Therefore, even accounting for a system capable of an infinite displacement, the number of points entering an infinitesimally small neighbourhood of the null state is limited to the set displacements outside it. Consequently, the number of system trajectories entering the neighbourhood is decreased by one every time it is increased in dimension. 
 
 To formalise what explained in the previous paragraph, let's consider a circumference in phase space ($\tilde{x},\dot{x}$) and increase its radius using unit step ($\Delta i=1$) to $\epsilon$ (i.e., $i\in[0,\epsilon]\subset\mathbb{N}$). The point-wise dimensionality for semicircle in divergence is:
 \begin{equation*}
     .5\sum_{i=1}^\epsilon\epsilon=.5\epsilon\sum_{i=1}^\epsilon1=.5\epsilon^2
 \end{equation*}
The point-wise dimensionality for the semicircle of convergence has to account for the decreased number of available trajectories for every increase leading to:
\begin{equation}
     \label{eq:dimconv}
     .5\sum_{i=1}^\epsilon\epsilon-i=.5\epsilon\sum_{i=1}^\epsilon\left(1-\cfrac{i}{\epsilon}\right)=.5\epsilon^{2-\gamma}
 \end{equation}
where $\gamma\in(0,~1]\subset\mathbb{R}$ describe reduction in dimensionality. The value of gamma depends on the maximum dimension of the neighbourhood ($\epsilon$) and the chosen discretisation ($\Delta i$), which asymptotically converges to zero when $(\Delta i\rightarrow0,\epsilon\rightarrow\infty)$. This can be easily verified by rewriting \autoref{eq:dimconv} as:  
\begin{equation*}
    .5\epsilon\sum_{i=1}^\epsilon\left(1-\cfrac{i}{\epsilon}\right)=5\epsilon\sum_{i=1}^\epsilon 1 - 5\epsilon\sum_{i=1}^\epsilon 1 -.5\sum_{i=1}^\epsilon i= .5\epsilon^2 -.5\sum_{i=1}^\epsilon i
\end{equation*}
and studying its limits for $\epsilon\rightarrow\infty$:
\begin{equation*}
    \lim_{\epsilon\rightarrow\infty} .5\epsilon^2 -.5\sum_{i=1}^\epsilon i = .5(\infty^2-\infty)=\infty^2
\end{equation*}
The total point-wise dimensionality of the Cantor set is obtained by adding the semicircle together.
\begin{equation}
    \label{eq:dim}
\begin{array}{rl}
     N_{(\tilde{x}=0,~\dot{x}=0)}\propto& 0.5\epsilon^2+0.5\epsilon\sum_{i=1}^\epsilon \left(1-\cfrac{i}{\epsilon}\right)= (.5\epsilon^2+.5\epsilon^{2-\gamma})
\end{array}
\end{equation}
\end{proof}

The attractor dynamics is defined to redistribute the potential energy accumulated in the controller to asymptotically return to the desired end-effector state using a smooth trajectory. The proposed method is based on the assumption that the energy reservoir is the controller's nonlinear stiffness. Consequently, the energy that the robot can release into the environment is upper-bounded by the potential energy accumulated in the controller. The Fractal Attractor also limits the controller power $\dot{E}$; thus, guaranteeing global stability in fixed-base robots.

\begin{figure}[!tb]
\begin{center}
    \begin{minipage}[b]{\columnwidth}
	\centering
	\includegraphics[width=.7\columnwidth]{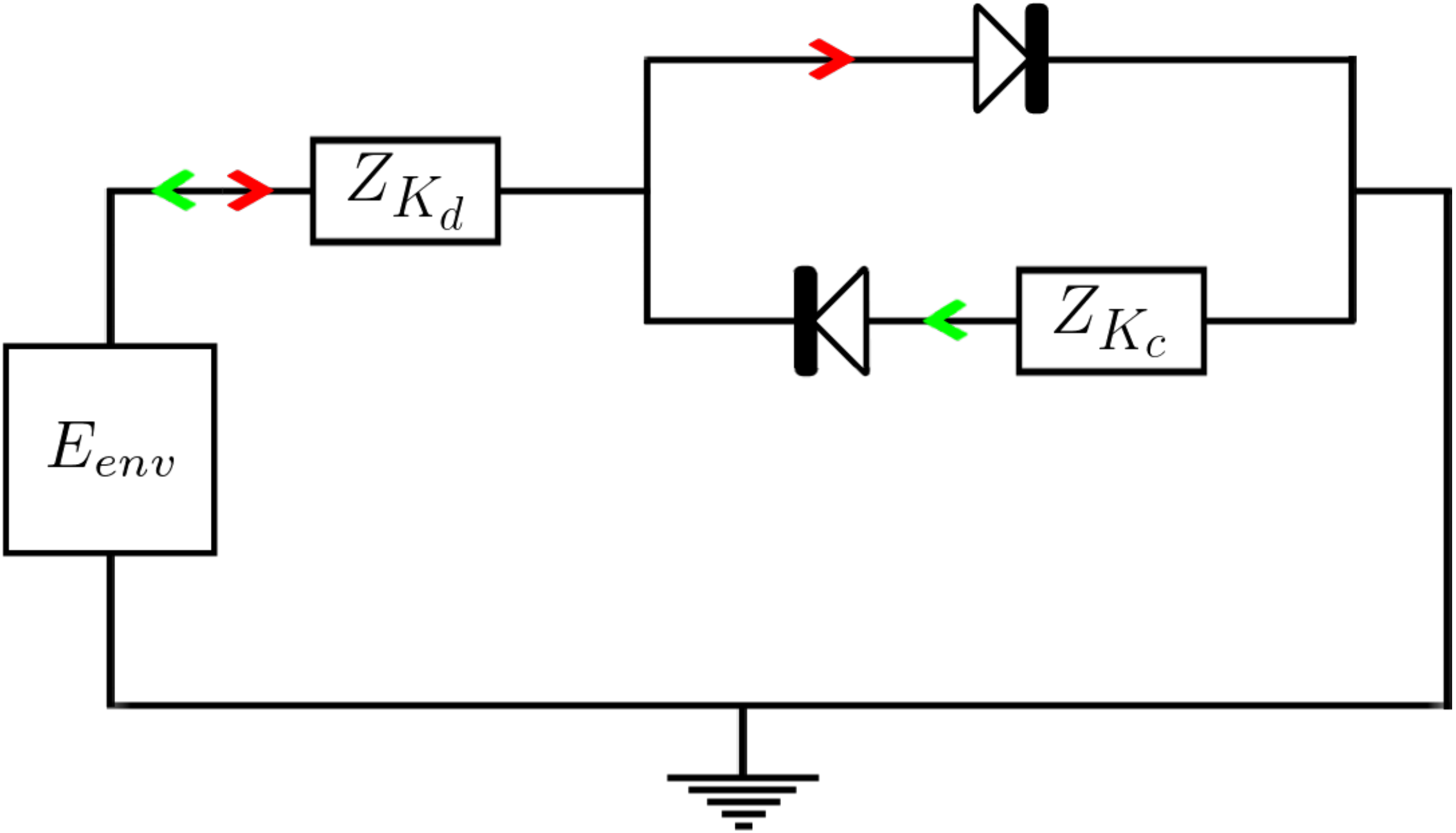}
	\subcaption{\centering\label{fig:4a} Energy Flow in Proposed Controller}
	\end{minipage}
	\hfill
	\begin{minipage}[b]{.7\columnwidth}
	\centering
	\includegraphics[width=\columnwidth]{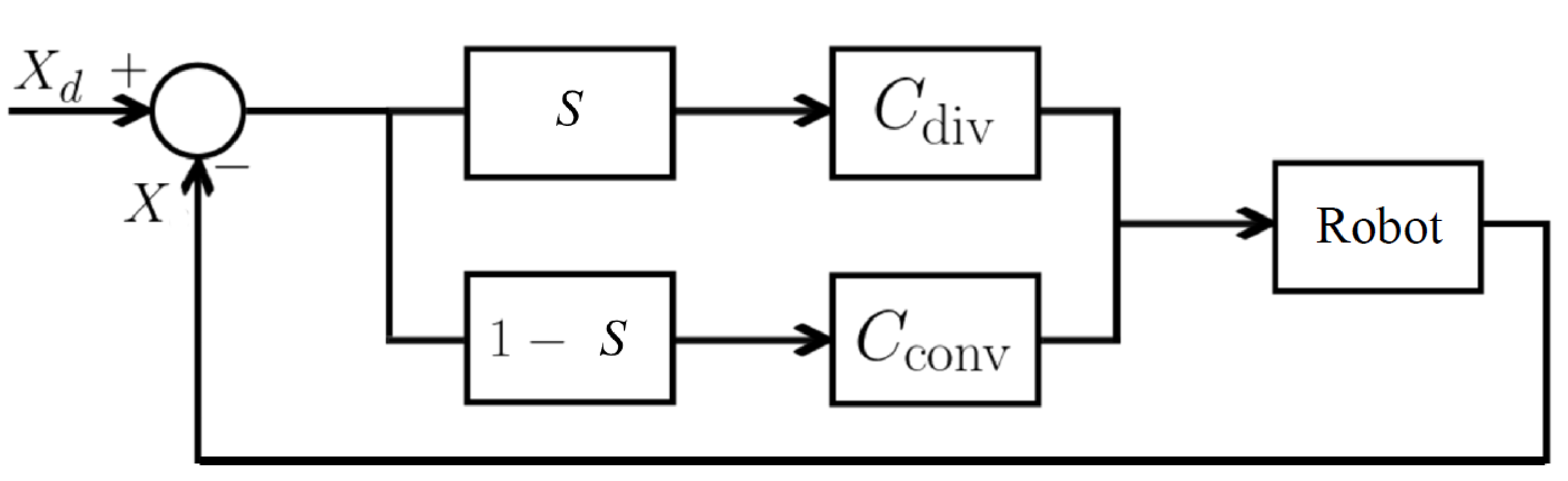}
    \subcaption{\centering\label{fig:4b}Block Diagram of Proposed}
	\end{minipage}
	\caption{ (a) The environment ($E_{env}$) introduces the energy in the robot during divergence, following the line marked by red arrows. This energy ($E_{\text{In}}$) is accumulated in the the stiffness component of the robot impedance ($Z_{K_d}$). During convergence, the robot releases the energy ($E_{\text{Out}}$) through the line marked by green arrows, engaging $Z_{K_c}$ in series to $Z_{K_d}$. (b) The activation of $C_{\text{div}}$ (impedance control in divergence phase) and $C_{\text{conv}}$ (passivity control in convergence phase) is performed by changing $S$, as described in \autoref{eq:3}.}
	\label{fig:4}
	\end{center}
\end{figure}

Energy redistribution is achieved by altering the controller's impedance when the end-effector starts to converge toward the desired pose. To do so an impedance $Z_{K_c}$ is added in series to the desired impedance $Z_{K_d}$, as shown in \autoref{fig:4a}. Subsequently, the original impedance  $Z_{K_d}$ is restored either when the system reaches the desired state or enters a new divergence phase. The impedance switch occurs at every divergence/ convergence boundary, as exemplified in \autoref{fig:4}.

\autoref{fig:3} describes the phase space for a 1-DoF attractor generated by the controller without damping. 
The divergence phases (quadrants $Qi$ and $Qiii$) have an impedance equal to $Z_{K_d}$.
The convergence phases (quadrants $Qii$ and $Qiv$) have an impedance equal to $Z_{K_d}+Z_{K_c}$. 
\autoref{fig:3c} shows a sample trajectory moving clockwise that branches in quadrant $Qiv$ into an ideal behaviour and a perturbed behaviour.
    
The switching behaviour is implemented using a hard-switch ($s=0,1$) and is based on the architecture shown \autoref{fig:4} and proposed by \cite{SwitchingArchitecture}.
In our case, the switch is activated by the following condition (independently for each task-space DoF):
\begin{equation} \label{eq:3}
    \left\{	
     \begin{array}{ll}
	{s} = 1		& \text{if~} \dot{x}=0~ \lor ~ \text{sgn}(\tilde{x}) = \text{sgn}(\dot{x})\\\\
	{s} = 0	& \text{o/w}
	\end{array} 
	\right.
\end{equation}
\noindent where $s =1$  denotes the divergent phase and ${s} =0$  denotes the convergent phase for the considered DoF. When ${s} =1$ the stiffness of the impedance controller $C_{\text{div}}$ is equal to $k_d(\tilde{x})$ as in Equation~\ref{eq:2}. 
At the moment $s$ becomes $0$, the impedance controller $C_{\text{conv}}$ is derived by applying the conservation of energy principle to ensure the controller's passivity. The updated impedance redistributes the accumulated energy, using half for acceleration towards the equilibrium point and the other half for deceleration once the position error has been halved. Thus, it produces balanced acceleration and deceleration phases and reduces the peak control torques. The mathematical formulation of the stiffness profile for the controller $C_{\text{conv}}$ is obtained by applying the conservation of energy between the energy accumulated in the robot stiffness at the beginning of convergence ($E_\text{In}$) and the energy that will be released during convergence ($E_\text{Out}$) for each task-space DoF:

\begin{equation} \label{eq:4}
	\left.
	\begin{array}{l}
	E_{\text{In}}=\int k_{d}\left(\tilde{x}\right)\tilde{x}~d\tilde{x}\\\\
	E_{\text{Out}}=2\cfrac{k^{'}_{\text{total}}\tilde{x}_{\text{mid}}^2}{2}=k^{'}_{\text{total}}\tilde{x}_{\text{mid}}^2\\\\
	{\text{Passivity} \implies E_{\text{In}}\geq\displaystyle{E_{\text{Out}}}}\\\\
    k^{'}_{\text{total}} =\left( \cfrac{1}{\tilde{x}^2_{\text{mid}}}\right) E_{\text{In}}=\left( \cfrac{4}{\tilde{x}^2_{\text{max}}}\right) E_{\text{In}}
	\end{array}
	\right.
\end{equation}
where $\tilde{x}_{\text{max}}$ is the displacement reached at the moment of switch from $s=1$ to $s=0$, and $\tilde{x}_{\text{mid}}$ is the midpoint between $\tilde{x}_{\text{max}}$ and $x_d$.
Note that $k^{'}_{\text{total}}$ in  \autoref{eq:4} assumes the spring's equilibrium point is moved to $\tilde{x}_{\text{mid}}$ during convergence. Rather we maintain $x_d$ as the equilibrium point by introducing the following nonlinear stiffness:
\begin{equation} \label{eq:5}
\left.
\begin{array}{l}
    k_{\text{total}}(\tilde{x}) =  k^{'}_{\text{total}} \left(\cfrac{0.5\tilde{x}_{\text{max}} -\tilde{x}}{\tilde{x}}\right) 
\end{array}
\right.
\end{equation}
which satisfies the condition:
\begin{equation*}
    k_{\text{total}}\tilde{x} =k^{'}_{\text{total}}\tilde{x}_{\text{mid}}~, ~ \forall~ \tilde{x}~ \in (0,~ \tilde{x}_{\text{max}})
\end{equation*}
\autoref{eq:4} imposes the controller passivity, which is a sufficient condition for stability. Note that the controller does not require damping in order to dissipate energy.

Finally, the FIC is added together with dynamics compensation and a null space controller term to obtain the robot control torques ($\tau_\text{ctrl}$) as described in \autoref{alg:1}; where $q$ is the joint configuration vector, $\dot{q}$ is the joint velocity vector, $M(q)$ is inertia matrix, $J(q)$ is the jacobian, $G(q)$ is the gravity compensation, and t$ C(q,\dot{q})$ is the Coriolis matrix.    
\begin{algorithm} 
1\text{~~}\textbf{for }$ i \in [1,6] \subset \mathbb{N}$ \textbf{do}\\
2\text{~~\quad}\textbf{if} diverging from $X_{d,i}$ (${S}_{i} = 1$) \textbf{do} \\
3\text{~~\qquad}$K_{ii}$ = ${K_d}_{,ii}$ (\autoref{eq:2})\\
4\text{~~\quad}\textbf{else do}\\
5\text{~~\qquad}$K_{ii}$ = ${K_{\text{total},ii}}$  (\autoref{eq:5})\\
6\text{~~\quad}\textbf{end}\\
7\text{~~}\textbf{end}\\
8\text{~~}$\Lambda (q) = {(JM^{-1}J^{T})}^{-1}$\\
9\text{~~}$\bar{J}^T = {(JM^{-1}J^{T})}^{-1}JM^{-1}$\\
10\text{~}$W_{\text{ctrl}} = K\tilde{X} + D\dot{\tilde{X}}+ \Lambda (q)( JM^{-1}C(q,\dot{q})\dot{q} - \dot{J}\dot{q})+G(q)$\\
11\text{~}${\tau}_{\text{ctrl}} = J^T W_{\text{ctrl}} + (I - J^{T}{\bar{J}}^T) {\tau}_{\text{null}}$ 
\caption{Fractal Impedance Control}
\label{alg:1}
\end{algorithm}

\subsection{Lyapunov Stability Analysis} \label{sec:stabilityAnalysis}
The proposed controller is characterised by a non-smooth piece-wise energy manifold with a time-invariant topology, as described by \autoref{eq:4}. 
In other words, the system dynamics is deterministic and changing the controller gains online does not affect either the manifold regularity or the fact that the system energy flow is always converging to the null state $(\tilde{x}=0,~\dot{x}=0)$.
This simplifies the stability analysis as it implies that changing gains are just an alteration of the initial condition. 
Furthermore, the intrinsic damping can be assumed to be zero without loss of generality in the proposed controller. 
Let us now consider the controller's autonomous dynamics for a monodimensional system:
		\begin{equation}
	\label{eq:6}
	   \left\{
	   \begin{array}{ll}
	    \lambda(x)\ddot{x} +\dot{\lambda}(x)\dot{x}  + k_d\tilde{x}=0 & \text{ Divergence}\\
	    \lambda(x)\ddot{x} +\dot{\lambda}(x)\dot{x} + k_{\text{total}}\tilde{x}=0 & \text{ Convergence}
	    \end{array}\right.
	\end{equation}
	
\noindent {where $\lambda (x)$ is the task-space inertia. $\dot{\lambda}(x)\dot{x}$ are the Coriolis and Centrifugal forces \cite{Sciavicco}. A valid Lyapunov's candidate is:}
	\begin{equation}
	\label{eq:7}
	\begin{cases}
\displaystyle{	V_{\text{div}}= \frac{\dot{x}^{T}\lambda(x) \dot{x}}{2} +\int k_{d}\tilde{x}~d\tilde{x}} \\ 
	\displaystyle{V_{\text{con}}= \frac{\dot{x}^T\lambda (x)\dot{x}}{2} +\frac{\tilde{x}^T k_{\text{total}}\tilde{x}}{2}+\frac{\int_{0}^{\tilde{x}_{\text{max}}} k_{d}\tilde{x}~d\tilde{x}}{2}} 
	\end{cases}
	\end{equation}
\noindent {Time derivative of $V$ is:}
	\begin{equation}
	\label{eq:8}
	   \begin{cases}
	    \dot{V}_\text{div}=(\lambda(x)\ddot{x} +\dot{\lambda}(x)\dot{x}  + k_d\tilde{x})\dot{x}=0\\
	    \dot{V}_\text{con}=(\lambda(x)\ddot{x} +\dot{\lambda}(x)\dot{x} + k_{\text{total}}\tilde{x})\dot{x}=0
	    \end{cases}
	\end{equation}
 \autoref{eq:7} and \autoref{eq:8} prove that the two pieces that compose our manifold are stable. 
 However,  Lyapunov's stability for non-smooth systems also requires to verify that the candidate is a Lipschitz function, which implies $V$ exists and has a bounded finite derivative also at the transition point.
 To verify the continuity condition for the Lyapunov function, the limits of the two continuous functions at the switching conditions, occurring for $\dot{x}=0$, should be the same value. 
 When the switching occurs for $\tilde{x}\ne0$, this is guaranteed by  \autoref{eq:5}. On the other hand, such equality needs to be verified for $\tilde{x}=0$: 
	\begin{equation}
	\label{eq:9}
	\begin{cases}
	\displaystyle{\lim_{\substack{\tilde{x}\to0 \\ \dot{x}\to0}}{V_{\text{div}}}=\int k_{d}\tilde{x}~d\tilde{x}=0}\\
    \displaystyle{\lim_{\substack{\tilde{x}\to0 \\ \dot{x}\to0}}{V_{\text{conv}}}=\frac{\tilde{x}^T k_{\text{total}}\tilde{x}}{2}+\int_{0}^{\tilde{x}_{\text{max}}} k_{d}\tilde{x}~d\tilde{x}=}
    \\~~~~~~=\displaystyle{\left(\int_{0}^{\tilde{x}_{\text{max}}} k_{d}\tilde{x}- k_{d}\tilde{x}~d\tilde{x}\right)=0}
	\end{cases}
	\end{equation}
To verify that V is radially unbounded integrating the energy associated with $K_\text{d}$ in \autoref{eq:2}.
\begin{equation}
\label{eq:10}
E=\left\{\begin{array}{lc}
 \cfrac{e^{\left(\beta\tilde{x}\right)^2}}{2\beta}+\cfrac{ k_{\text{const}}\tilde{x}^2}{2},  &\text{}\lvert\tilde{x}\rvert \leq x_{B}\\\\
   w_\text{max}\tilde{x}+\cfrac{e^{\left(\beta\tilde{x}_\text{B}\right)^2}}{2\beta} +\cfrac{k_{\text{const}}\tilde{x}_\text{B}^2}{2}  , & \text{o/w} 
 \end{array}\right.
\end{equation}
The controller energy is radially unbounded (i.e., $E \to \infty \text{ if } \lvert x \rvert \to \infty$), and consequentially also $V$ (\autoref {eq:7}) is radially unbounded.
\begin{equation*}
\displaystyle{\lim_{\tilde{x}_{\text{max}}\to \infty} V_{\text{div}}(x)=E(x)= \int_{0}^{\infty} k_{d}\tilde{x}~d\tilde{x}=\infty}
\end{equation*}
Furthermore, the $\dot{V}=0$ in both the transition states $(\tilde{x}_{\text{max}})$ and $(\dot{x}=0)$. Therefore, the controller respects all the required stability conditions.

\subsection{FIC Intrinsic Robustness to Low-Bandwidth Feedback}
The FIC generates a conservative field with asymptotically stable autonomous trajectories. Thus, the attractor energy is path-independent. As a consequence, the proposed controller is intrinsically robust to discretisation, model errors and noise. On the other hand, this is not valid for virtual energy tank methods which rely on an integrator to track the energy exchanged by the controller to achieve passivity. To verify our claim we analyse the computation of energy exchanged by the FIC and a task-space mass-spring-damper impedance controller (i.e., PID in velocity). Consequently, the proportional, integral, and derivative terms are the damping, the stiffness, and the inertia, respectively.

\begin{figure}[!tb]
    \centering
	\includegraphics[width=.75\columnwidth,trim=5.5cm 10cm 5.5cm 10.4cm,clip]{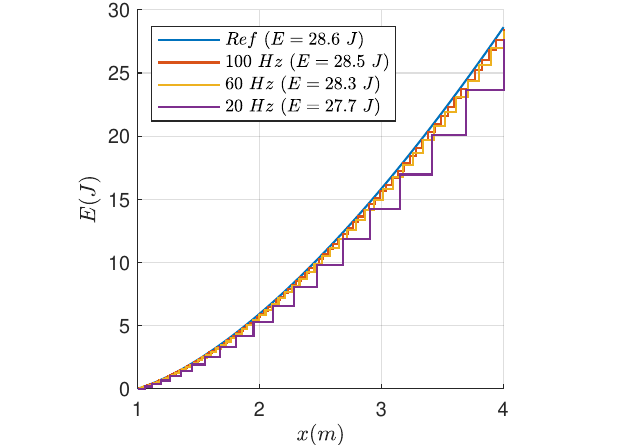}
    \caption{$\Delta E_\text{IC}$ for a system ($k_\text{D}=1 \SI{}{\kilo\gram}$, $k_\text{P}=1\SI{}{\newton\per\meter}$ ) tracking the trajectory $x(t)=t^3+t^2+t$ for $1~\SI{}{\second}$ at different sampling rates are compared against the expected reference value computed at a frequency of $10\SI{}{\kilo\hertz}$. The system energy ($\text{E}$) computed with the different samplings of the same signal shows an error that increases with the reduction of the bandwidth. The drift of the energy values reveals that integration errors become relevant for low-bandwidth controllers even for simple ideal dynamics (absence of both model errors and noise).}
   \label{fig:5}
\end{figure}

 The mechanical work of a mono-dimensional FIC is equal to the difference in potential energy between the two states (i.e., A and B):
	\begin{equation}
		\label{eq:11}
		\begin{array}{ll}
		 \Delta E_\text{FIC}&=\displaystyle{\int_{A}^{B}f(\tilde{x})dS=\int_{A}^{B}f(\tilde{x})\dot{\tilde{x}}dt}\\\\
		 &=\displaystyle{\int_{A}^{B}f(\tilde{x})dx}=E(B)-E(A)
		\end{array}
	\end{equation}
\noindent where E is the energy of associated with the controller state, and reported in \autoref{eq:10}.
As a consequence, the integral is path independent, and the evaluation of the controller energy is unaffected by the sampling time in discrete systems. Let's now consider the mechanical work of an impedance controller using constant gains: 
\begin{equation}
	\label{eq:12}
	\begin{array}{ll}
		\displaystyle{\Delta E_\text{IC}}&=\displaystyle{\int_{A}^{B}f(\tilde{x},\dot{\tilde{x}},\ddot{\tilde{x}})dS}=\displaystyle{\int_{A}^{B}f_\text{D}(\ddot{\tilde{x}},\dot{\tilde{x}})dS+}\\\\	&\displaystyle{+\int_{A}^{B}f_\text{P}(\dot{\tilde{x}})dS+\int_{A}^{B}f_\text{I}(\tilde{x})dS=}\\\\
		&=\displaystyle{k_\text{D}\int_{0}^{t_\text{f}}\ddot{\tilde{x}}\dot{\tilde{x}}dt+k_\text{P}\int_{0}^{t_\text{f}}\dot{\tilde{x}}\dot{\tilde{x}}dt+k_\text{I}\left.\frac{\tilde{x}}{2}\right\rvert_A^B}
		\end{array}
\end{equation}
where $t_\text{f}$ is the final time of the trajectory. $k_\text{D}$, $k_\text{P}$ and $k_\text{I}$ are inertia, damping and stiffness of the controller. As $k_\text{I}$ is a conservative component, it is path independent and unaffected by switching to discrete-time systems. However, this is not valid for both inertia and damping, which are both trajectory dependant. Thus, the calculation of mechanical work for an impedance controller depends on the chosen discretion and will always be an approximation of the effective work in \autoref{eq:12}. For example, let's consider the non-conservative components of the impedance controller's mechanical work and apply a zero-order hold time discretisation. 
\begin{equation}
\label{eq:13}
\begin{array}{ll}
	\displaystyle{\Delta E_\text{IC}^-}&=\displaystyle{k_\text{D}\sum_{A}^{B}\ddot{\tilde{x}}\Delta\tilde{x}+k_\text{P}\sum_{A}^{B}\dot{\tilde{x}}\Delta\tilde{x}}=\\ &=\displaystyle{k_\text{D}\sum_{t=0}^{t_\text{f}}\ddot{\tilde{x}}(t)(\tilde{x}(t+\Delta t)- \tilde{x}(t))\Delta t+}\\ &	+\displaystyle{k_\text{P}\sum_{t=0}^{t_\text{f}}\dot{\tilde{x}}\left(t\right)\left(\tilde{x}\left(t+\Delta t\right)-\tilde{x}\left(t\right)\right)\Delta t}
\end{array}
\end{equation}
It shall be noted that we have assumed for simplicity that the time needed to move between $A\to B$ is a multiple of the sampling time for all the selected $\Delta t$. The estimation of the energy can be improved with state estimators. However, their accuracy will be always be compromised by a low sampling frequency (\autoref{fig:5}), model errors and noise. 

\section{Experimental Validation} \label{sec:sim_exp_validation}
The experiments are designed to validate if the proposed controller retains the desirable characteristics and performance of impedance controllers whilst reducing some of its shortcomings. In other words, we would like to retain the interaction robustness of impedance controllers while being able to adjust online the trade-off between compliance and tracking accuracy. 
	
A 7-DoF Panda robot manipulator from Franka Emika is used for the experiments where the controller damping is always set to zero. 
During contact experiments, a  PTFE sheet is attached to the end-effector to reduce sliding friction. An ATI Gamma SI-130-10 force/torque (F/T) sensor, mounted at the end-effector of the Panda, records the  interaction forces but is not used in the controller. 
	
\subsection{End-Effector Pose Tracking Before Controller Calibration}\label{sec:desried_endEff_pose_tracking}

Diagonal gain matrices are used for the constant stiffness gains in \autoref{eq:2}. The values of $k_{\text{const}}$ are set to $\SI{150}{\newton\per\meter}$ for linear degrees of freedom (DoF) and $\SI{5}{\newton\meter\per\radian}$ for the angular DoF.
We also test the proposed impedance profile on a traditional impedance controller (\autoref{eq:1}) by turning off the Fractal Attractor (stiffness set to \autoref{eq:2} during both convergence and divergence). However, the high stiffness values may render the manipulator unsafe and trigger the Panda safety mechanisms, as shown in the attached video. Therefore, it is not possible to run a comparison with a traditional impedance controller using the nonlinear stiffness of \autoref{eq:2}. 

\subsubsection{Static Target}
The initial configuration of the end-effector is set to $X_{d} = [0.5 \ 0 \ 0.5 \ -\pi \ 0 \ 0]^T$ (\SI{}{\meter~or~\radian}), and the end-effector is randomly perturbed by a human operator 15 times. The robot's recovery behaviour is recorded.
The initial virtual boundaries, controlled by tuning the nonlinear stiffness profile in \autoref{eq:2}, are as follows: $X_{B,1:3} = \SI{0.05}{\meter}$ and $X_{B,4:6} = \SI{0.1746}{\radian}$, unless differently stated. 
The mean $\bar{\tilde{X}}$, standard deviation $\sigma$, and the RMSE of the end-effector error are used to evaluate the system accuracy. The mean and standard deviation of the convergence time after a perturbation are computed to evaluate the recovery behaviour of the controller.

\subsubsection{Trajectory Tracking} \label{traj_tracking}
We evaluate the robustness of the proposed method to track a desired trajectory when the end-effector is perturbed. 
The desired periodic trajectory, on y-axis, has amplitude $\pm \SI{0.25}{\meter}$ and period $T = \SI{40}{\second}$.
Fifteen manually generated perturbations of the end-effector are used for this evaluation.
The mean and the standard deviation of RMSE for $\tilde{X}$ are computed to evaluate the controller tracking performance. 
The convergence time mean and standard deviation are also measured to evaluate the performance of the controller after each perturbation.

\subsection{End-Effector Pose Tracking and Forward Force Control after Controller Calibration} \label{sec:Calibration}
Despite the controller allowing global stability, each physical system has marginal stability due to its limited finite properties, such as  power and band-pass limits. Thus, within the context of this method, controller calibration refers to the process of identifying the upper-bounds of the controller's parameters that can be applied without exceeding the physical properties of the robot within its work-space. 
We set the damping to zero and then evaluate the maximum force that can be exerted without losing stability. 
It is worth noting that the value of force would not precisely correspond to the amount of force exerted on the end-effector due to the presence of model errors. 
The authors would like to remark that these calibrations are both system and impedance profile specific.
    
The following is the calibration process used to identify the maximum force that can be applied for a given boundary when the damping is set to zero:
\begin{enumerate}[i)]
		\item Setting $K_{\text{const},ii} = 0$ ($\forall i \in [1,6] \subset \mathbb N$)
		\item Setting the initial maximum allowed wrench at the virtual boundaries by taking the maximum payload of the robot into account: $W_{\text{max},1:3} = \SI{30} {\newton}$ and $W_{\text{max},4:6} = \SI{20}{\newton\meter}$
		\item Perturbing the end-effector of the robot and reducing the size of the virtual constraints ($X_{{B,i}}$), until the robot starts to oscillate \label{CAL3}
		\item Reducing $W_{\text{max},i}$ and keeping the value for $X_{{B,1:6}}$ before the oscillation \label{CAL4}
		\item Repeating steps \autoref{CAL3} and \autoref{CAL4} until: $X_{{B,1:3}} = \SI{0.001}{\meter}$ and $X_{{B,4:6}} = \SI{0.0174}{\radian}$
\end{enumerate}
	
The controller is tested again after calibration, and trajectory tracking performance is evaluated with the same method used for the trajectory tracking experiment above. 
Furthermore, the stability and the accuracy of the force interaction are evaluated using the force/torque sensor mounted at the end-effector. 
The force interaction has been limited to the z-axis for the scope of this paper. 
The following experiments are carried out during this phase:
\begin{enumerate}[1)]	
        \item \textit{Static \& Trajectory Tracking:} The previous experiments are repeated after calibration. 
        \item \textit{Online Virtual Boundary Adjustment:} The human operator randomly perturbs the robot in static as its virtual constraint size of $X_{B,1:3}$ is decreased online. 
        $X_{B,1:3}$ reduces from $\SI{0.20}{\meter}$ by $\SI{0.0025}{\meter}$ every second until it reaches $\SI{0.001}{\meter}$. 
        During the interaction, the manipulator stiffness automatically increases from initial value of $k_{d,2} = \SI{150}{\newton\per\meter}$ at the beginning of the experiment where $X_{B,1:3} = \SI{0.20}{\meter}$ to $k_{d,2} = \SI{1250}{\newton\per\meter}$ at the end due to the change in the virtual constraints. 
        \item \textit{Circular Trajectory Tracking:} The trajectories on the $xy$-, $xz$- and $yz$-planes are tracked to evaluate the accuracy of the controller across multiple directions. 
        The starting pose is $X_{d} = [0.4 \ 0 \ 0.85 \ 0 \ 0 \ 0]^T$, the trajectory execution period $T=\SI{40}{\second}$, and its $\text{radius} = \SI{0.1}{\meter}$.
        The tracking performance has been evaluated initially for $K_{\text{const}} = \mathbf{0}$, then setting $K_{\text{const},ii} = \SI{150}{\newton\per\meter}$ for the linear degrees of freedom and $K_{\text{const},ii} = \SI{5}{\newton \meter\per\radian}$ for the angular degrees of freedom. 
        \item \textit{Interaction With Objects:} The robot makes contact with Box \#1 shown in \autoref{fig:8a} and gradually pushes down to exert the maximum allowed force in the direction of z-axis.
        The force evolution over time is recorded with the force/torque sensor and compared with the data from the forward model.
        The virtual constraints are $X_{B,1:3} = \SI{0.1}{\meter}$ and $X_{B,4:6} = 0.174$~\\~$6~\si{\radian}$.~~~~~~~~~~
\end{enumerate}

\subsection{Robustness to Low-Bandwidth Feedback}
The FIC's robustness to low-bandwidth feedback is evaluated on a static pose and trajectory tracking tasks along the lateral direction (y-axis). In both cases, the controller behaviour is analysed in both unperturbed and perturbed conditions.  Low-bandwidth feedback is emulated by applying a zero-order hold to the robot's sensory feedback signals. The perturbations used are 10 pushes manually applied at the end-effector along the vertical directions (z-axis). The RMSE and recovery time of the perturbations are calculated for experiments executed with feedback frequencies of \SI{20}{\hertz}, \SI{100}{\hertz} and \SI{1000}{\hertz}. 

All the low-bandwidth experiments were conducted setting $X_{B}=[7.5~7.5~7.5~1.047~1.047~1.047]^T$ \SI{}{\centi\meter~or~\radian}, $K_\text{const}=\mathrm{diag}(100,~100,~100,~5,~5,~5)$  \si{\newton\per\meter~or~\newton\meter\per\radian}, and a passive damping $D=\mathrm{diag}(2.5,~2.5,~2.5,~1.25,~1.25,~1.25)$  \si{\newton\second\per\meter} or \si{\newton\meter\second\per\radian}. The passive damping was introduced to experimentally verify that the stability of the system is not compromised as long as there is no velocity tracking. Lastly, it shall be noted that during this experiment, the Franka Arm was wrapped in plastic covers due to a concurrent experiment involving sand.

\section{Results} \label{sec:RESULTS}
 The passivity of the controller is verified by checking the difference between the absorbed $E_{\text{in}}$ and energy released on the environment $E_{\text{rel}}$, computed as follows:
	\begin{equation}
	\begin{array}{lc}
	\displaystyle{E_{\text{in}} = \sum_{i=1}^{6} \int K_{d,ii}{\tilde{X}}_{i}~d\tilde{X}_{i}},& \text{Divergence \ \ } \\
	E_{\text{rel}} = \max \left(\cfrac{1}{2} \dot{X}^T \Lambda (q) \dot{X}\right), & \text{Convergence}
	\end{array}{}
	\label{eq:14}
	\end{equation}

\subsection{End-Effector Pose Tracking Before Controller Calibration}
The collected data for the static perturbations in \autoref{tab:2} indicate that the position error is constrained and consistent. 
The average convergence time over the 15 perturbations is $\SI{1.43}{} \pm \SI{0.047}{\second}$  and the difference between the absorbed and released energy in static is $E_{\text{rel}} - E_{\text{in}} = -0.011 \pm 0.003 \ \si{\joule}$. 
The trajectory tracking data are reported in \autoref{tab:2}. The average convergence time to the desired pose is $\SI{1.45}{} \pm \SI{0.047}{\second}$, and the energy exchanged is $-0.019 \pm 0.006 \ \si{\joule}$.
  
  \begin{figure}[!tb]
\centering
	\begin{minipage}[b]{0.4\columnwidth}
    	\centering
	    \includegraphics[width=\columnwidth]{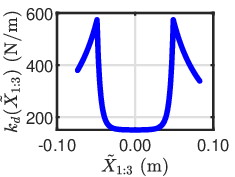}
	    \subcaption{\centering \label{fig:6a}}
	\end{minipage}
	\begin{minipage}[b]{0.4\columnwidth}
	    \centering
	    \includegraphics[width=\columnwidth]{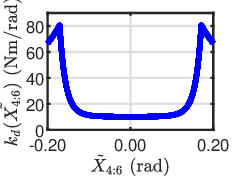}
	    \subcaption{\centering \label{fig:6b}}
    \end{minipage}
    \caption{Stiffness Profile w.r.t $X_{B,1:3} = \SI{0.05}{\meter}$ and $X_{B,4:6} = \SI{0.1746}{\radian}$  - (a) Variable stiffness w.r.t position error (${\tilde{X}}_{1:3}$), (b) Variable stiffness w.r.t orientation error (${\tilde{X}}_{4:6}$).}
	\label{fig:6}
\end{figure}
\subsection{Controller Calibration}
\autoref{tab:1} represents the results of the controller calibration and it shows how the value of $W_{\text{max}}$ changes as $X_B$ varies.
A preliminary analysis of the stiffness profile obtained with the Franka Panda has been performed to verify that the proper impedance profile has been implemented in the robot. The results are shown in  \autoref{fig:6}, which are congruent with the theoretical profiles shown in \autoref{fig:2}.
	\begin{table*}[!htb]
	\caption{Controller Calibration w.r.t $W_{\text{max}}$ and $X_{B}$. The forces and torques are expressed in \si{\newton} and \si{\newton\meter}, respectively. The position are reported in \si{\centi\meter}, and the angles are expressed in \si{d\radian}.} 
	\begin{center}
		\begin{tabular}{llllll} 
		\toprule
		\multicolumn{6}{c}{Linear} \\
		$W_{\text{max},x}$ &  $X_{{B,x}}$ & $W_{\text{max},y}$ &  $X_{{B,y}}$ & $W_{\text{max},z}$ &  $X_{{B,z}}$  \\ 
		\midrule
		30 & $\geq$ \SI{3}{} & 30 &  $\geq$ \SI{9}{} & 30 &  $\geq$ \SI{8}{}\\ 
		15 & $[$ \SI{1}{}, \SI{3}{} $)$ & 25 & $[$ \SI{5}{}, \SI{8}{} $)$ & 28 & $[$ \SI{5}{}, \SI{8}{} $)$ \\ 
		12.5 & $[$ 0, \SI{1}{} $)$ & 22.5 & $[$ \SI{4}{}, \SI{5}{} $)$ & 26 & $[$ \SI{4}{}, \SI{5}{} $)$ \\ 
		// &  // & 20 & $[$ \SI{3}{}, \SI{4}{} $)$ & 24 & $[$ \SI{3}{}, \SI{4}{} $)$ \\ 
		// &  // & 18 & $[$ \SI{2}{}, \SI{3}{} $)$ & 20 & $[$ \SI{2}{}, \SI{3}{} $)$ \\ 
		// &  // & 12.5 & $[$ \SI{0.7}{}, \SI{2}{} $)$ & 12.5 & $[$ \SI{0.7}{}, \SI{2}{} $)$ \\
		// &  // & 10 & $[$ \SI{0.6}{}, \SI{0.7}{} $)$ & 10 & $[$ \SI{0.6}{}, \SI{0.7}{} $)$ \\ 
		// &  // &  6 & $[$ \SI{0.4}{}, \SI{0.6}{} $)$ & 6 & $[$ \SI{0.4}{}, \SI{0.6}{} $)$ \\ 
		// &  // &  5 & $[$ \SI{0.3}{}, \SI{0.4}{} $)$ & 5 & $[$ \SI{0.3}{}, \SI{0.4}{} $)$ \\ 
		// &  // &  4 & $[$ \SI{0.2}{}, \SI{0.3}{} $)$ & 4 & $[$ \SI{0.2}{}, \SI{0.3}{} $)$ \\ 
		// &  // &  3 & $[$ \SI{0.1}{}, \SI{0.2}{} $)$ & 3 & $[$ \SI{0.1}{}, \SI{0.2}{} $)$ \\
		// &  // &  2 & $[$ 0, \SI{0.1}{} $)$ & 2 & $[$ 0, \SI{0.1}{} $)$ \\ \bottomrule
		\end{tabular}
		\begin{tabular}{llllll}
		\\
		\toprule
		\multicolumn{6}{c}{Angular} \\ 
		$W_{\text{max},\phi_x}$ &  $X_{{B,\phi_x}}$ & $W_{\text{max},\phi_y}$ &  $X_{{B,\phi_y}}$ & $W_{\text{max},\phi_z}$ &  $X_{{B,\phi_z}}$  \\ 
		\midrule
		20 & $[$ \SI{1.04}{}, $10\pi$ $]$  & 20 &  $[$ \SI{1.04}{}, $10\pi$ $]$  & 20 & $[$ \SI{2.61}{}, $10\pi$ $]$ \\
		15 & $[$ \SI{0.87}{}, \SI{1.04}{} $)$ & 15 & $[$ \SI{0.87}{}, \SI{1.04}{} $)$ & 15 & $[$ \SI{0.69}{}, \SI{2.61}{} $)$ \\ 
		5 & $[$ \SI{0.17}{}, \SI{0.87}{} $)$ & 5 & $[$ \SI{0.17}{}, \SI{0.87}{} $)$ & 5 & $[$ \SI{0.17}{}, \SI{0.69}{} $)$ \\ 
		\bottomrule
		\end{tabular}
	\end{center}
	\label{tab:1}
	\end{table*}
\addtolength{\belowcaptionskip}{0pt}
\renewcommand{\arraystretch}{1.4}
\addtolength{\belowcaptionskip}{0pt}

\subsection{End-Effector Pose Tracking and Forward Force Control after Controller Calibration} 
\textit{Static \& Trajectory Tracking:} The data recorded for the static poses experiment after calibration are reported in \autoref{tab:2}. They indicate that there is a reduction of the pose error after a perturbation. The static position errors get $7$ times smaller after calibration. Meanwhile, the orientation errors get $3$ times smaller after calibration. The calibration also reduces the RMSE of the tracking errors of a factor between $2.5$ and $3.5$.  The convergence time to the desired pose is $\SI{1.38}{} \pm \SI{0.044}{\second}$ and the difference of absorbed and release energy is $ -0.033 \pm 0.010 \ \si{\joule}$. These results are also confirmed by the trajectory tracking experiment data reported in \autoref{tab:2}. 
The convergence time is $\SI{1.42}{} \pm \SI{0.045}{\second}$ and the difference between the absorbed and released energy is $-0.048 \pm 0.013 \ \si{\joule}$ during trajectory tracking. 

\begin{sidewaystable}[!htb]
    \small\sf\centering
	\caption{Static (S) and Trajectory Tracking (TT) Errors recorded before (BC) and after (AC) calibration.  WP indicates experiments with perturbations. NP refers to experiments without perturbations. $\bar{\tilde{X}}$ is the mean, and $\sigma$ is the standard deviation.}
	\begin{center}
	\begin{tabular}{lllllll} 	
	\toprule
	& S-BC & S-AC  & TT-BC-NP  & TT-AC-NP& TT-BC-WP& TT-AC-WP \\ 
	&  $\bar{\tilde{X}} \pm \sigma$ & $\bar{\tilde{X}} \pm \sigma$&  RMSE & RMSE & $\bar{\tilde{X}} \pm \sigma$ &  $\bar{\tilde{X}} \pm \sigma$  \\ 
	\midrule
	$\tilde{x} (\SI{}{\centi\meter})$   & $  \SI{3.52}{}\pm \SI{0.11}{} $ & $  \SI{0.43}{}\pm \SI{0.01}{} $ &  $ \SI{3.28}{} $ & $ \SI{1.14}{} $ & $ \SI{3.32}{} \pm \SI{1.06}{} $ & $ \SI{1.24}{} \pm \SI{0.40}{} $  \\ 
	$\tilde{y} (\SI{}{\centi\meter})$   & $  \SI{3.48}{}\pm \SI{0.11}{} $ & $\SI{0.32}{}\pm \SI{0.01}{} $ & $ \SI{3.46}{} $ & $ \SI{1.08}{} $&  $ \SI{3.54}{} \pm \SI{1.14}{} $ & $ \SI{1.20}{} \pm \SI{0.38}{} $  \\
	$\tilde{z} (\SI{}{\centi\meter})$	  & $  \SI{3.83}{}\pm \SI{0.12}{} $ & $  \SI{0.35}{}\pm \SI{0.01}{} $ &  $ \SI{3.67}{} $ & $ \SI{1.17}{} $ &$ \SI{3.73}{} \pm \SI{1.20}{} $ &  $ \SI{1.26}{} \pm \SI{0.41}{} $  \\  
	$\tilde{\phi}_x (\SI{}{\centi\radian})$ & $\SI{3.23}{}\pm \SI{0.10}{}  $ & $\SI{1.14}{}\pm \SI{0.03}{}  $ &  $ \SI{4.11}{} $ & $ \SI{1.29}{} $& $ \SI{4.13}{} \pm \SI{1.33}{} $ &  $ \SI{3.38}{} \pm \SI{1.09}{} $  \\  
	$\tilde{\phi}_y (\SI{}{\centi\radian})$ & $\SI{2.72}{}\pm \SI{0.08}{}  $ & $\SI{1.24}{}\pm \SI{0.04}{}  $ &   $ \SI{4.36}{} $& $ \SI{1.18}{} $  & $ \SI{4.48}{} \pm \SI{1.44}{} $ &  $ \SI{3.22}{} \pm \SI{1.03}{} $ \\
	$\tilde{\phi}_z (\SI{}{\centi\radian})$ & $\SI{3.17}{}\pm \SI{0.10}{} $ & $\SI{-1.38}{}\pm \SI{0.04}{} $ & $ \SI{4.69}{} $& $ \SI{1.38}{} $ &    $ \SI{4.74}{} \pm \SI{1.52}{} $ &  $ \SI{5.44}{} \pm \SI{1.10}{} $ \\
	\bottomrule
	\end{tabular}
	\end{center}
	\label{tab:2}
\end{sidewaystable}

\textit{Online Virtual Boundary Adjustment:} The stability is not affected by an online change of the virtual constraints while a user introduces random perturbations to the robot. The system remains passive over the trial with a difference between the absorbed and the released energy of $-0.136 \pm 0.005 \ \SI{}{\joule}$.	

\begin{figure}[!tb] 
    \centering
	\includegraphics[width=.7\columnwidth]{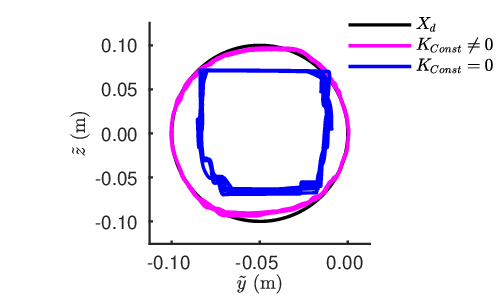}
	\caption{Circular trajectory tracking on yz-plane.}
	\label{fig:7}
\end{figure}

\textit{Circular Trajectory Tracking:} \autoref{tab:3} reports RSME tracking performances of end-effector pose on the xy-, xz- and yz-planes for $K_{\text{const}} \neq \mathbf{0}$. 
The best tracking performance is obtained with a boundary of $ 0.01 \SI{}{\metre}$ boundary when $K_{\text{const}} \ne \mathbf{0}$. 
The highest RMSE recorded is less than \SI{0.007}{\metre}. 
Sample circular trajectories for the yz-plane are provided in \autoref{fig:7}. 

\begin{table}[!htb]
    \caption{The mean of the RMSE measured for tracking a circular trajectory on 3 different planes. The values are expressed in $\SI{}{\milli\meter}$.}	
	\begin{center}
	\begin{tabular}{llll}
	\toprule
	\multicolumn{4}{c}{ $X_{B,1:3} = \SI{0.01}{(\meter)}$}\\
	$\text{plane}$ &  $\tilde{x}$ & $\tilde{y}$ &  $\tilde{z}$\\
	\midrule
	$\text{xy}$ & \SI{5.3}{} & \SI{2.2}{} &  \SI{3.1}{}\\ 
	$\text{xz}$ & \SI{5.2}{} & \SI{2.1}{} &  \SI{3.3}{}\\ 
	$\text{yz}$ & \SI{5.3}{} & \SI{2.4}{} &  \SI{3.2}{}\\ 
	\bottomrule
    \end{tabular}
    \begin{tabular}{llll}
	\toprule
	\multicolumn{4}{c}{$X_{B,1:3} = \SI{0.10}{(\meter)}$}\\
	$\text{plane}$ & $\tilde{x}$ & $\tilde{y}$ &  $\tilde{z}$\\
	\midrule
	$\text{xy}$ & \SI{23.6}{} & \SI{7.1}{} &  \SI{5.3}{} \\ 
	$\text{xz}$ & \SI{24.8}{} & \SI{4.3}{} &  \SI{5.6}{} \\ 
	$\text{yz}$ & \SI{18.4}{} & \SI{4.5}{} &  \SI{5.1}{} \\ 
	\bottomrule
    \end{tabular}
	\end{center}
	\label{tab:3}
\end{table}

\begin{figure}[!htb]
	\centering
	\begin{minipage}[b]{0.21\textwidth}
	\centering
	\includegraphics[width=\textwidth,height=4cm]{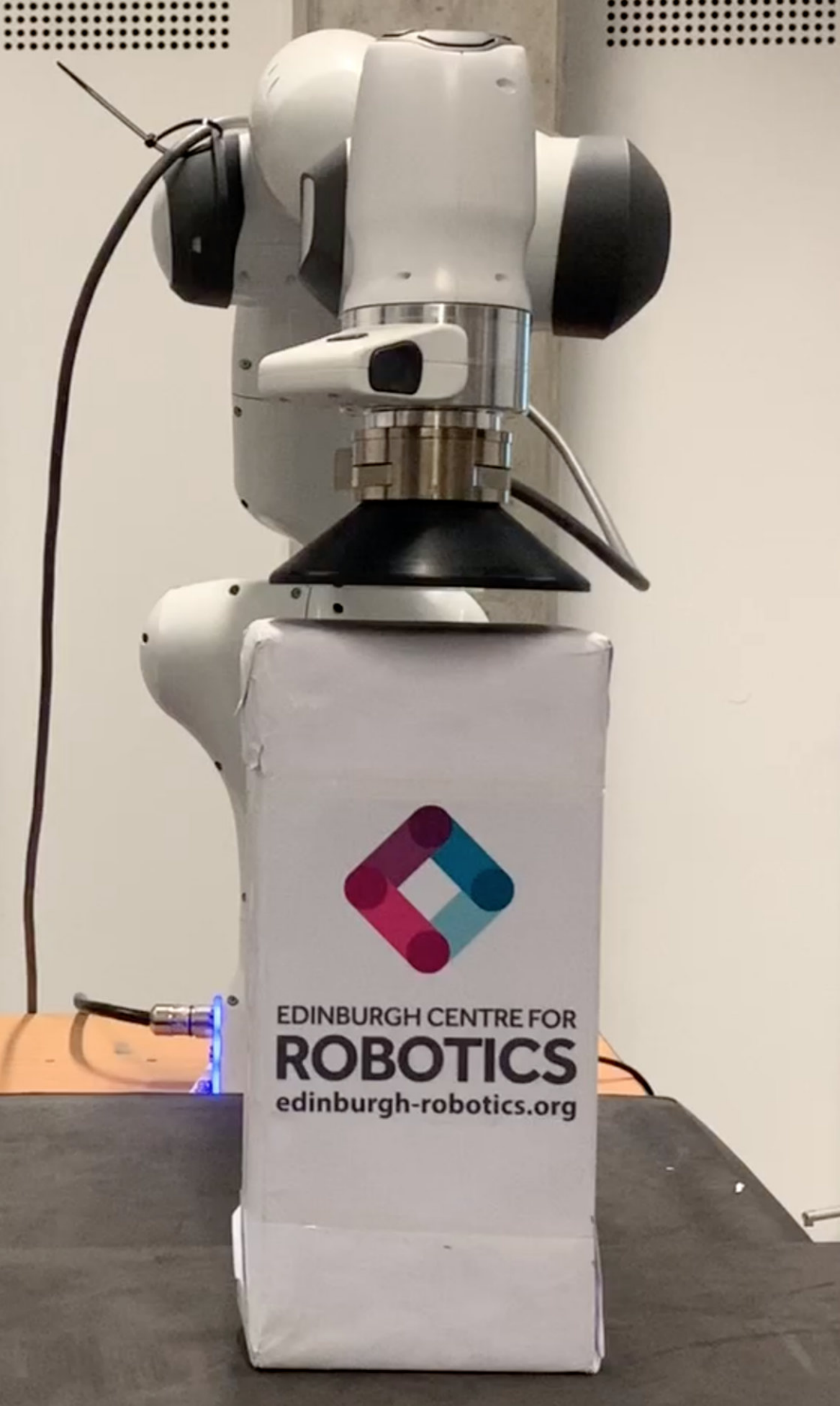}
	\subcaption*{\centering}
    \end{minipage}
		~
	\begin{minipage}[b]{0.21\textwidth}
	\centering
	\includegraphics[width=\textwidth,height=4cm]{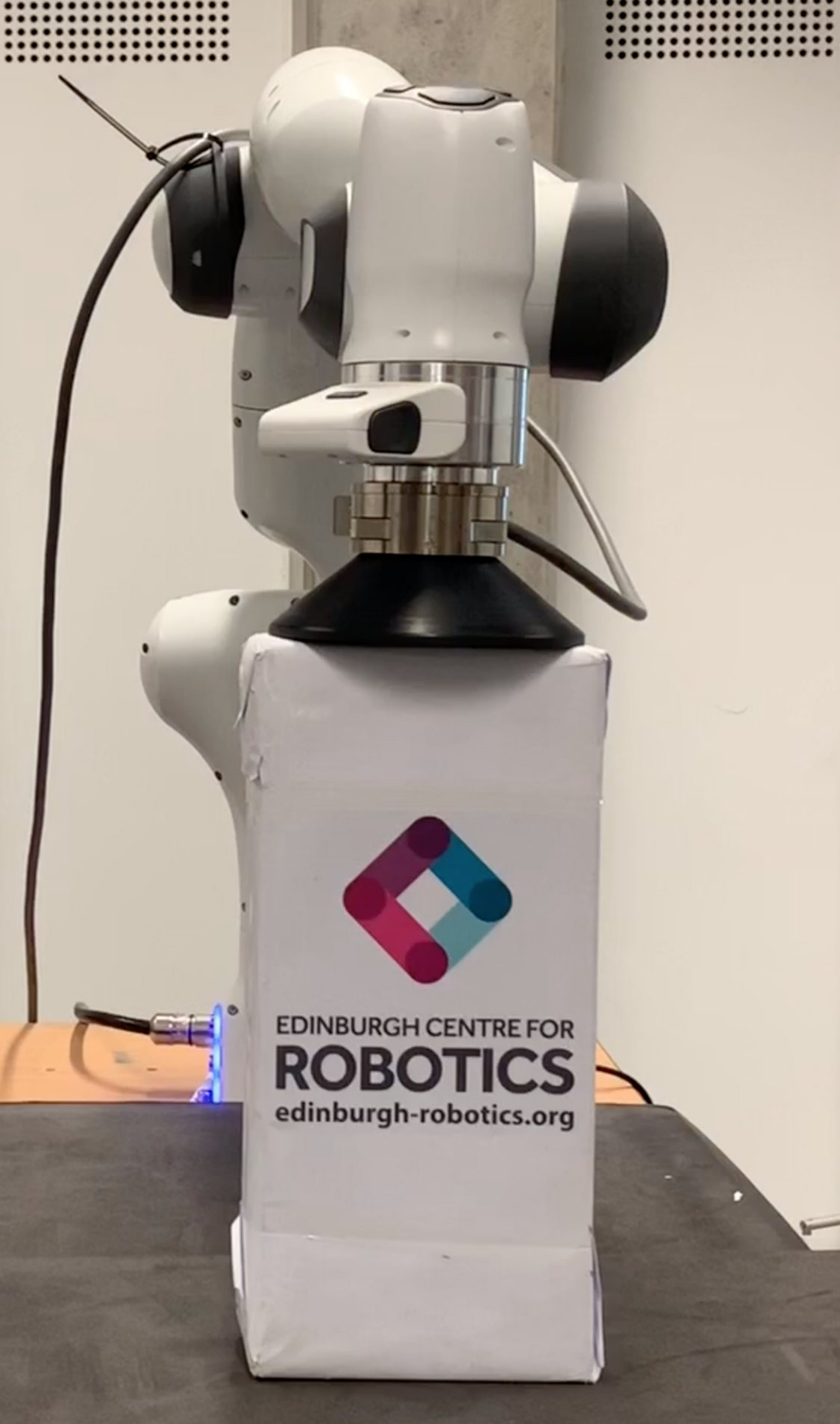}
	\subcaption{\centering \label{fig:8a}} 
	\end{minipage}
	~
	\begin{minipage}[b]{0.21\textwidth}
	\centering
	\includegraphics[width=\textwidth,height=4cm]{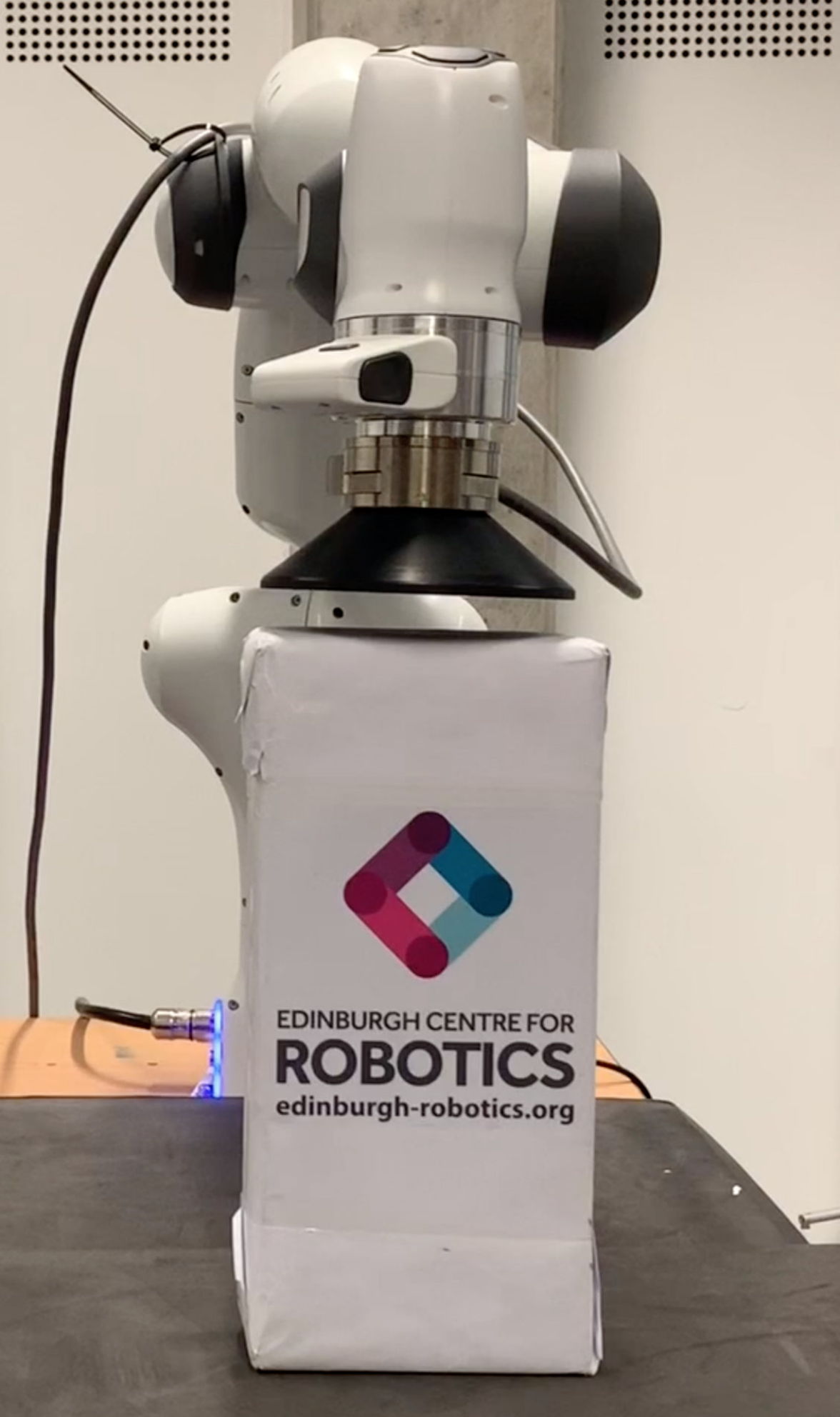}
	\subcaption*{\centering} 
	\end{minipage}
	~\\
	\begin{minipage}[b]{0.21\textwidth}
	\centering
	\includegraphics[width=\textwidth,height=4cm]{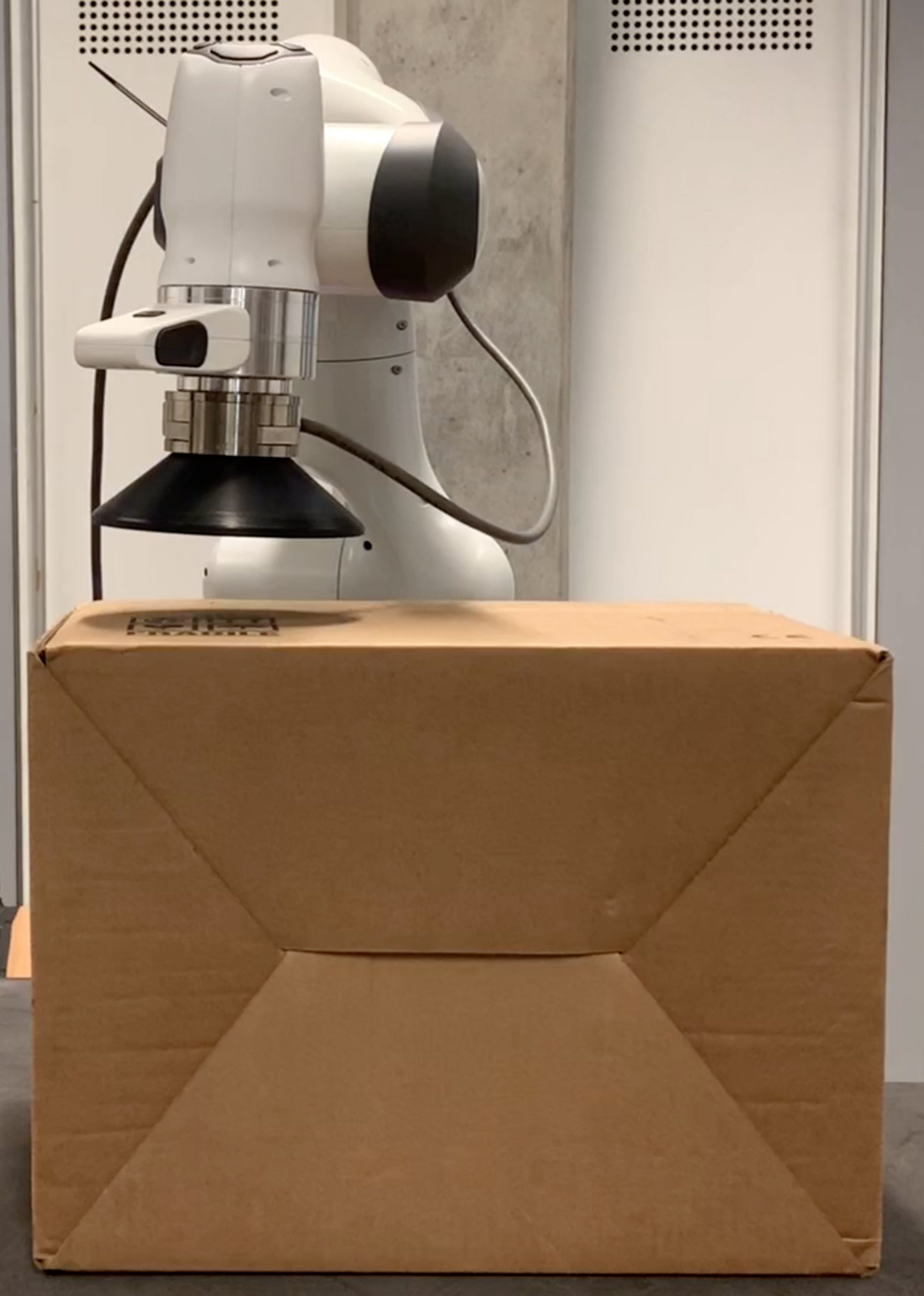}
	\subcaption*{\centering}
	\end{minipage}
	~
	\begin{minipage}[b]{0.21\textwidth}
	\centering
	\includegraphics[width=\textwidth,height=4cm]{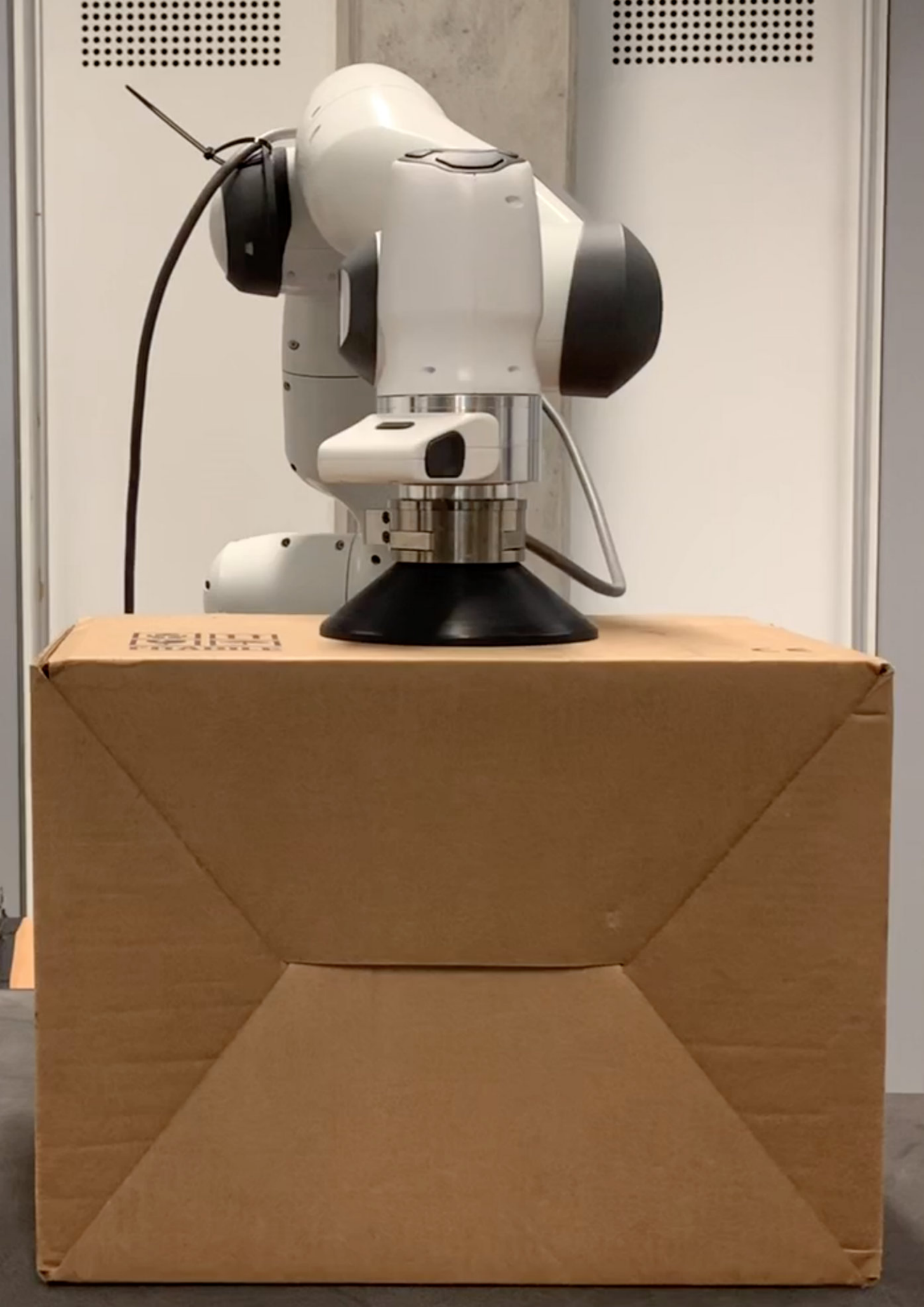}
	\subcaption{\centering\label{fig:8c}} 
	\end{minipage}
	~
	\begin{minipage}[b]{0.21\textwidth}
	\centering
	\includegraphics[width=\textwidth,height=4cm]{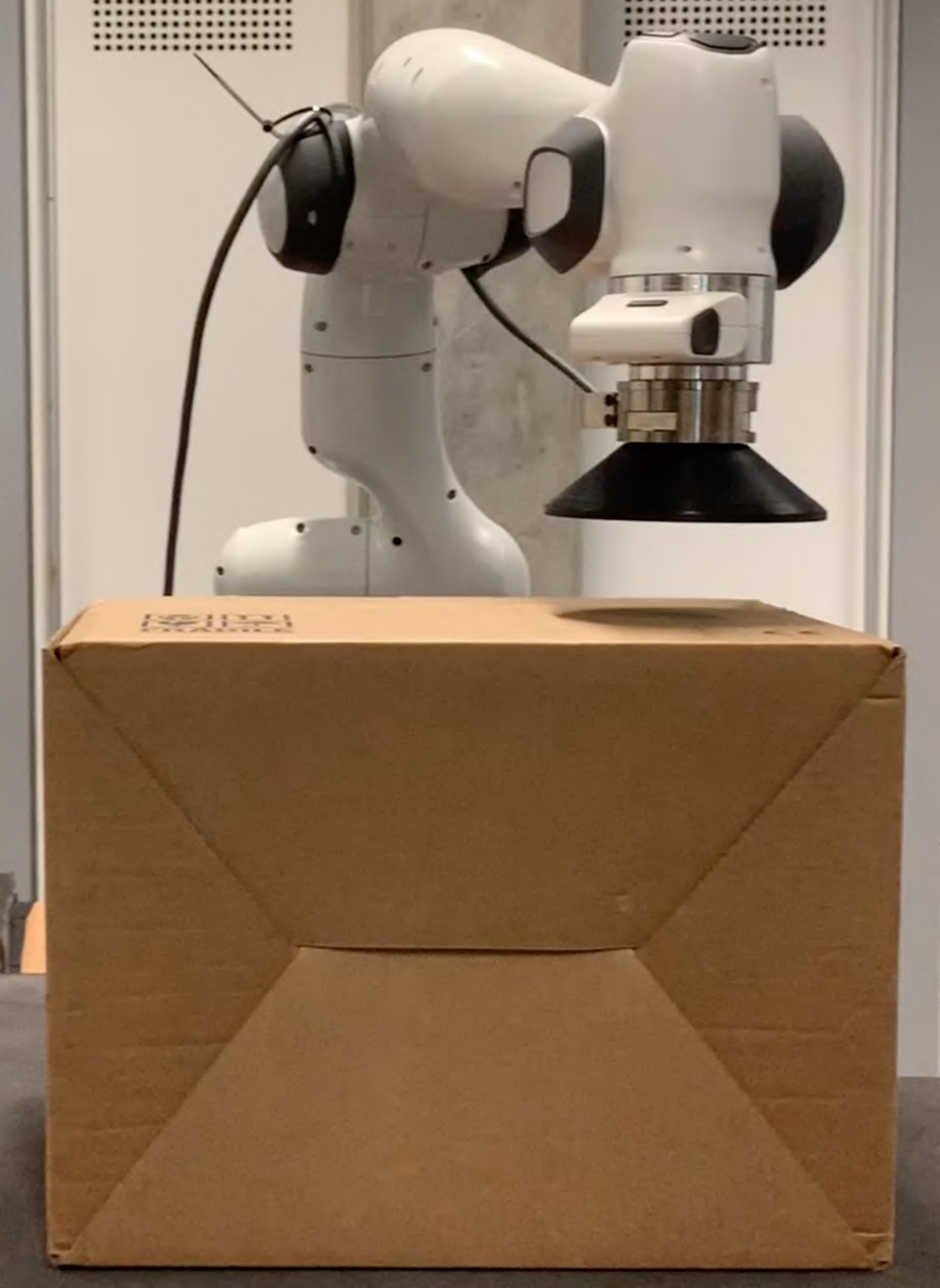}
	\subcaption*{\centering} 
	\end{minipage}
	~\\
	\begin{minipage}[b]{0.21\textwidth}
	\centering
	\includegraphics[width=\textwidth,height=4cm]{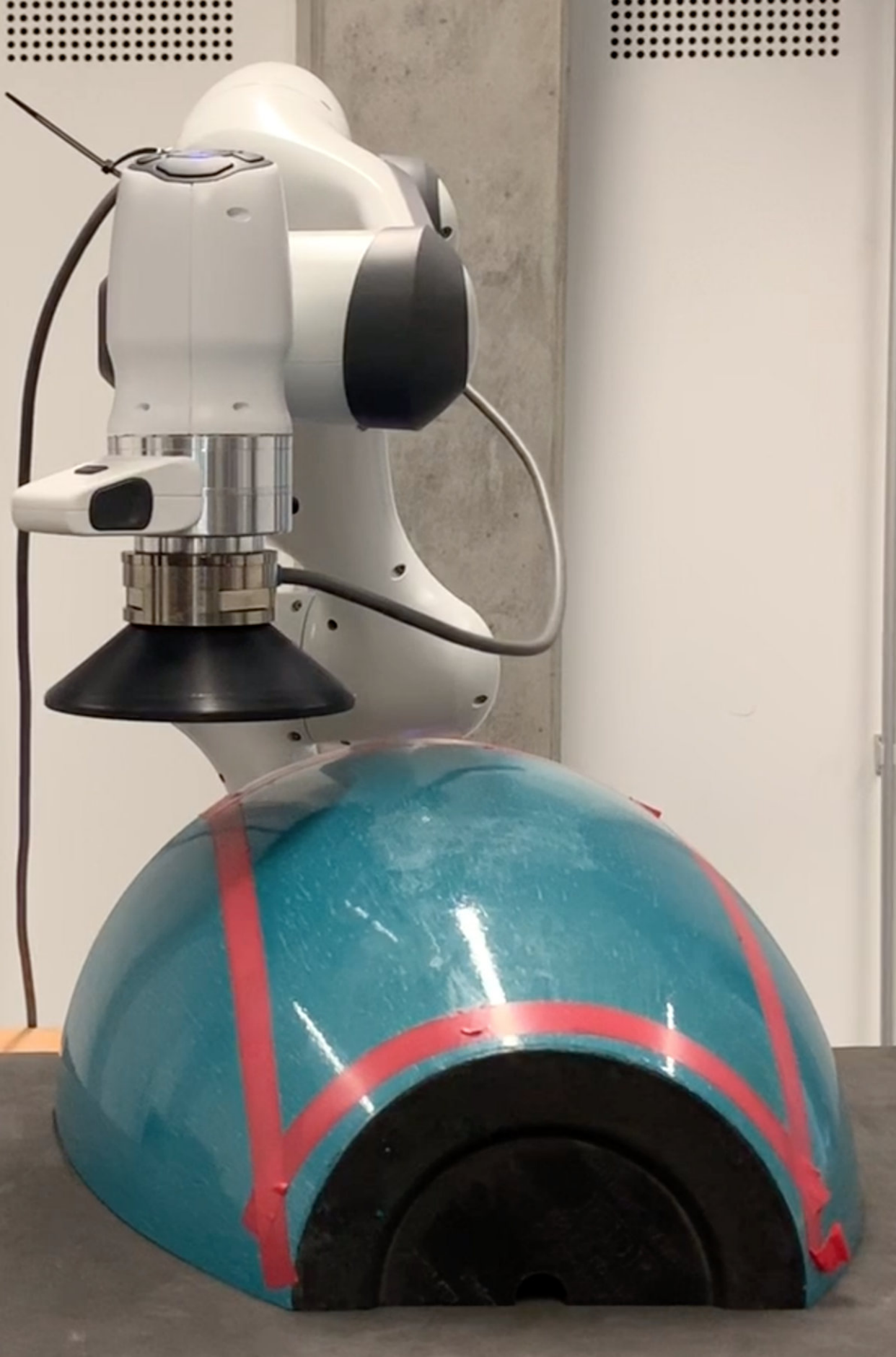}
	\subcaption*{\centering} 
	\end{minipage}
	~
	\begin{minipage}[b]{0.21\textwidth}
	\centering
	\includegraphics[width=\textwidth,height=4cm]{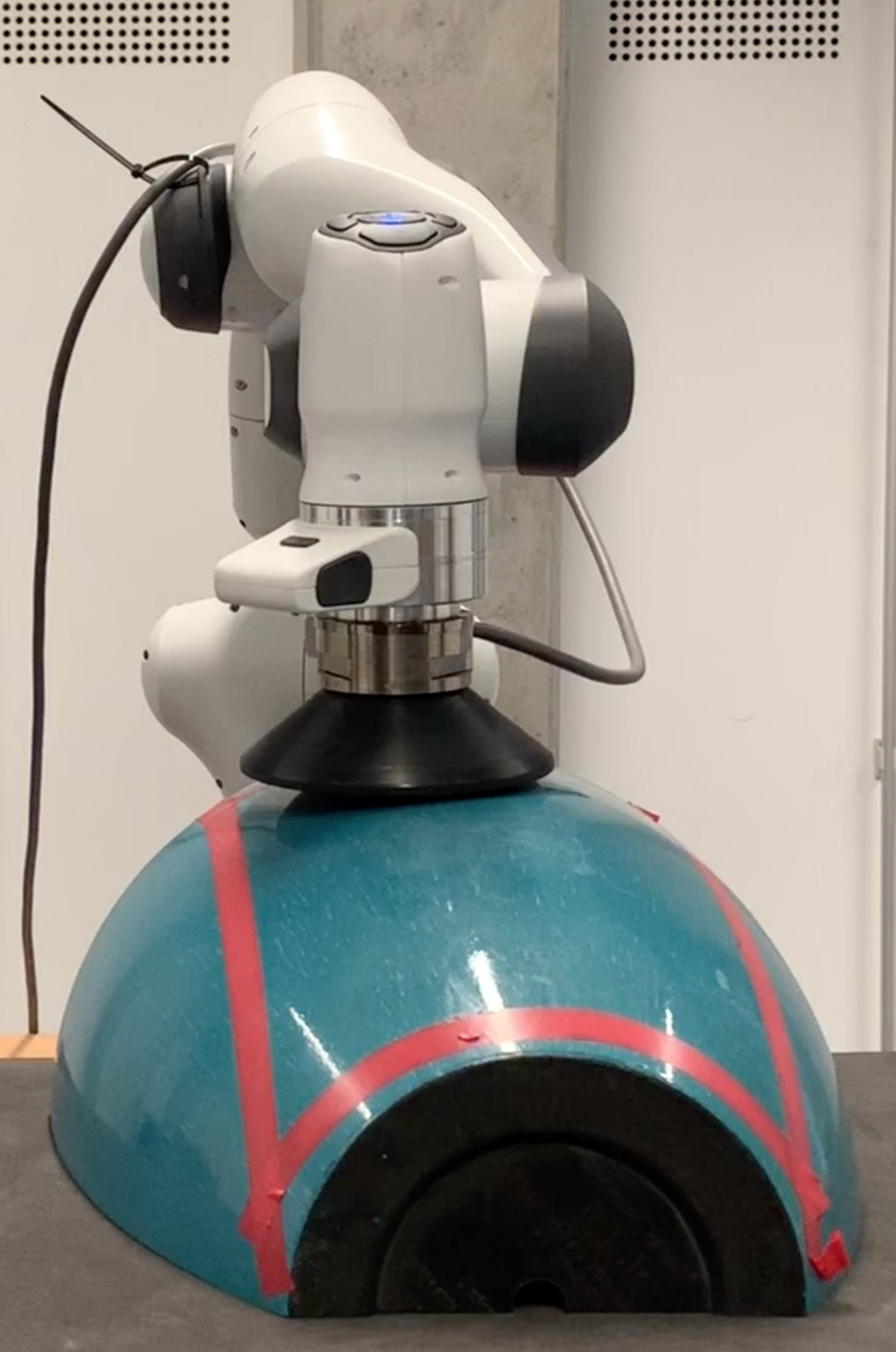}
	\subcaption{\centering\label{fig:8e}} 
	\end{minipage}
	~
	\begin{minipage}[b]{0.21\textwidth}
	\centering
	\includegraphics[width=\textwidth,height=4cm]{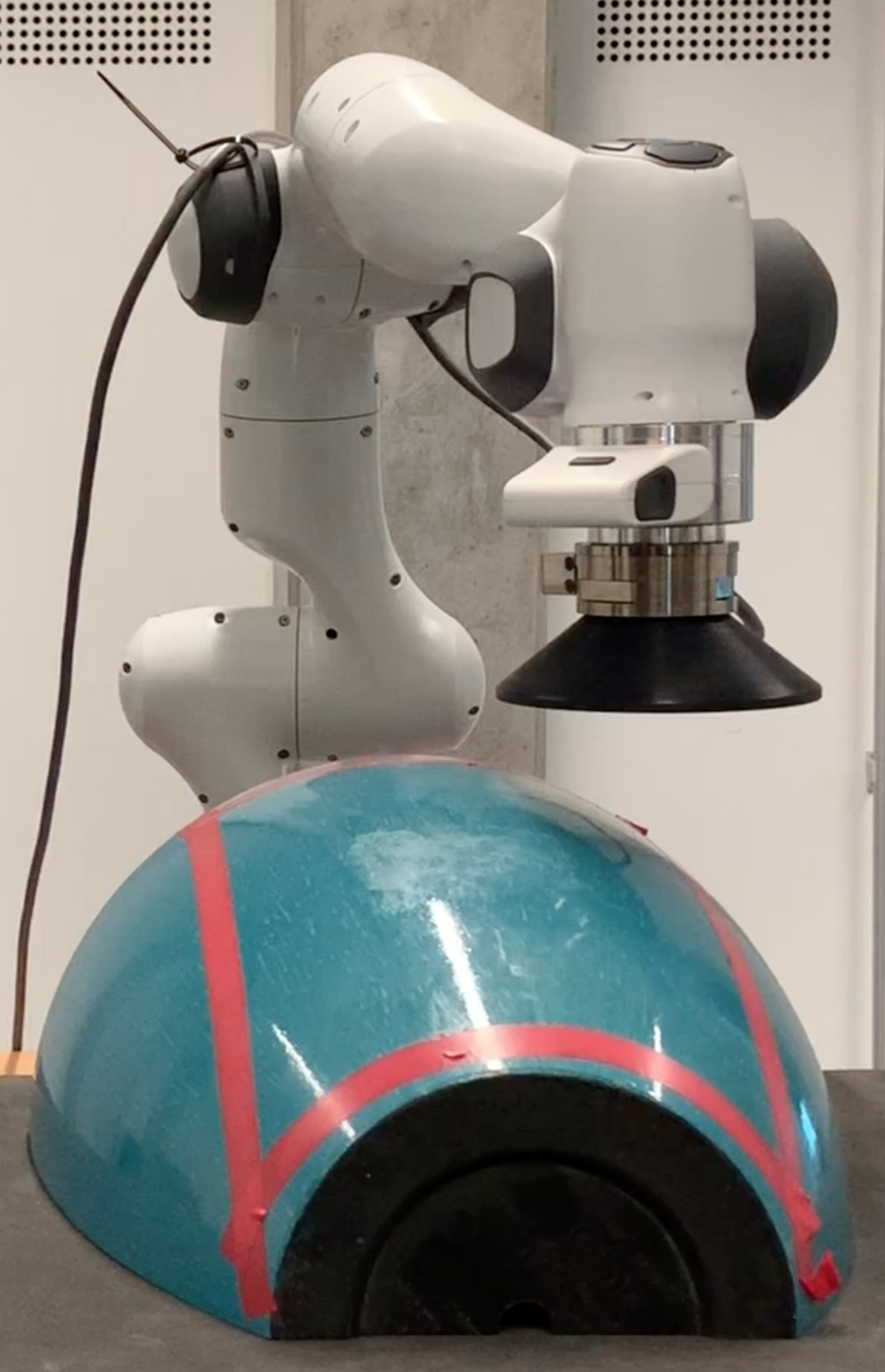}
	\subcaption*{\centering} 
	\end{minipage}
	\caption{Snapshots of the interaction experiments. (a) Vertical interaction with a cardboard box. (b) Interaction during tracking a sinusoidal trajectory (along the y-axis) while interacting with a cardboard box. (c) Tracking a sinusoidal trajectory along the y-axis while interacting with a curved surface.}
	\label{fig:8}
\end{figure}

\begin{figure}[!htb]
	\centering
	\begin{minipage}[b]{.7\columnwidth}
	\centering
	\includegraphics[width=\columnwidth]{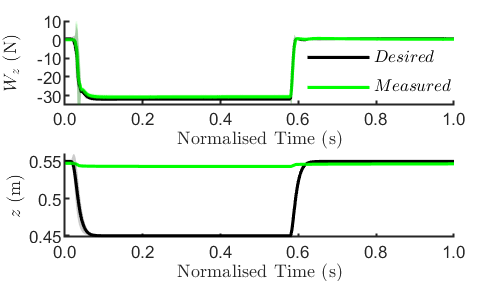}
	\subcaption{\centering\label{fig:8b}}
	\end{minipage}
	\begin{minipage}[b]{.7\columnwidth}
	\centering
	\includegraphics[width=\columnwidth]{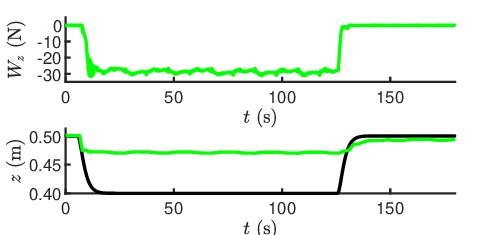}
	\subcaption{\centering\label{fig:8d}}
	\end{minipage}
	\begin{minipage}[b]{.7\columnwidth}
	\centering
	\includegraphics[width=\columnwidth]{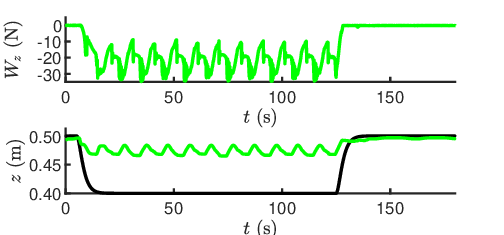}
	\subcaption{\centering\label{fig:8f}}
	\end{minipage}
	\caption{Data recorded in the interaction experiments. (a) Static interaction with a cardboard box normalised wrt time. The  shaded areas are the $90 \%$ prediction bounds for the recorded data. (b) Interaction with the cardboard box while tracking a sinusoidal trajectory along the y-axis. The average maximum force is $\SI{28.45}{\newton}$. (c) Interaction with the curved surface while tracking a sinusoidal trajectory along the y-axis. The average maximum force is $\SI{28.85}{\newton}$.}
	\label{fig:9}
\end{figure}

\textit{Interaction With Objects:} The steady state force recorded during the interaction with a box is $\SI{27.01}{\newton}$ which is the $95\%$ of the desired interaction force (\autoref{fig:8} and \autoref{fig:9}). 
	
\subsection{Robustness to Low-Bandwidth Feedback}
\begin{figure}[!tb]
\centering
\begin{subfigure}[b]{0.7\columnwidth}
	\centering
	\includegraphics[width=\columnwidth]{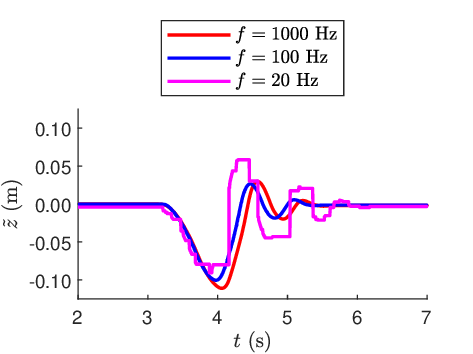}
	\caption{\label{fig:10a}}
\end{subfigure}
~~~
\begin{subfigure}[b]{0.7\columnwidth}
	\centering
	\includegraphics[width=\columnwidth]{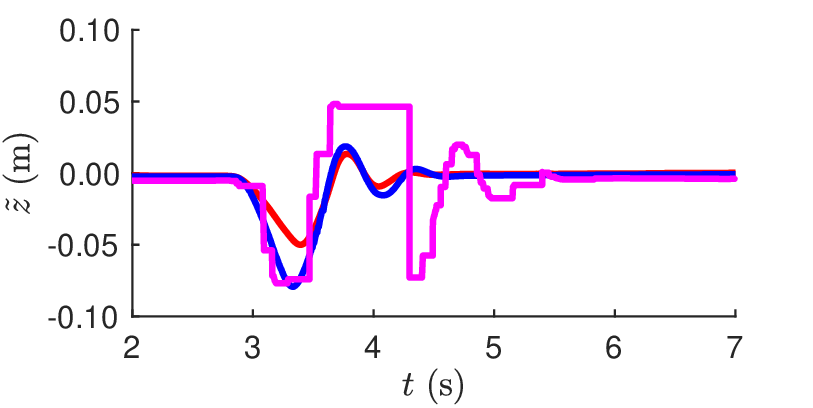}
    \caption{\label{fig:10b}}
\end{subfigure}
\caption{Comparison of one random interaction at the end-effector with the following feedback bandwidths: $\SI{1000}{\hertz}$, $\SI{100}{\hertz}$ and $\SI{20}{\hertz}$. (a) Static pose, (b) Trajectory Tracking. Reducing the feedback bandwidth reduces the tracking accuracy, but it does not compromise the task even when the system is perturbed.}
\label{fig:10}
\end{figure}

\begin{figure}[!tb]
\centering
\begin{subfigure}[b]{0.7\columnwidth}
	\centering
	\includegraphics[width=\columnwidth]{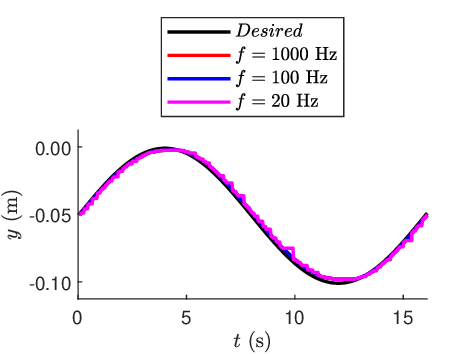}
	\caption{\label{fig:11a}}
\end{subfigure}
~~~
\begin{subfigure}[b]{0.7\columnwidth}
	\centering
	\includegraphics[width=\columnwidth]{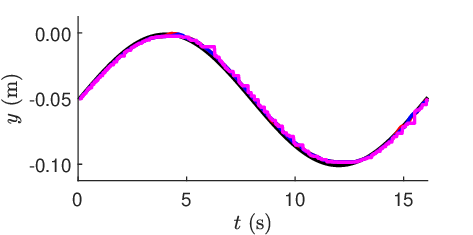}
    \caption{\label{fig:11b}}
\end{subfigure}
\caption{Comparison of the trajectory tracking along the $y$-axis with/without interaction on the end-effector with the following feedback bandwidths: $\SI{1000}{\hertz}$, $\SI{100}{\hertz}$ and $\SI{20}{\hertz}$. (a) Without interaction, and (b) With interaction (same experiment of \autoref{fig:10b}). A perturbation along the $z$-direction does not affect the task along the $y$-direction in any of the tested conditions.}
\label{fig:11}
\end{figure}

The controller remains stable after receiving an external perturbation even with a feedback bandwidth of $\SI{20}{\hertz}$ as shown in \autoref{fig:10} and \autoref{fig:11}. The data gathered in \autoref{tab:4} indicate that as the feedback frequency decreases the position error of $\tilde{x}$, $\tilde{y}$ and $\tilde{z}$ increases, but the tracking error at its highest peak is less than $\SI{5}{\milli\meter}$. The recovery time in the static task for the feedback frequencies of $\SI{1000}{\hertz}$, $\SI{100}{\hertz}$ and $\SI{20}{\hertz}$ are $\SI{1.33}{} \pm \SI{0.043}{\second}$, $\SI{1.45}{} \pm \SI{0.047}{\second}$ and $\SI{2.26}{} \pm \SI{0.074}{\second}$, respectively. The recovery time for the trajectory tracking task for the aforementioned feedback frequencies are: $\SI{1.37}{} \pm \SI{0.090}{\second}$, $\SI{1.51}{} \pm \SI{0.049}{\second}$ and $\SI{3.13}{} \pm \SI{0.103}{\second}$, respectively.

\begin{table}[!htb]
	\caption{Results of the Low-Bandwidth Feedback Experiments. S refers to static experiments. TT stands for Trajectory Tracking experiments. WP indicates experiments with perturbations. NP refers to experiments without perturbations. $\bar{\tilde{X}}$ is the mean, and $\sigma$ is the standard deviation.}
	\begin{center}
	\begin{tabular}{llll} 
	\toprule
    \multicolumn{4}{c}{f = \SI{1000}{\hertz}}\\
    & S & TT-NP & TT-WP  \\
	& $\bar{\tilde{X}} \pm \sigma$ & RMSE& $\bar{\tilde{X}} \pm {\sigma}$   \\ 
	\midrule
	$\tilde{x} (\SI{}{\milli\meter})$ & $\SI{1.4}{}\pm \SI{0.5}{} $  & $ \SI{3.1}{} $ & $ \SI{3.3}{} \pm \SI{1.1}{} $\\ 
	
	$\tilde{y} (\SI{}{\milli\meter})$   & $\SI{1.5}{}\pm \SI{0.5}{} $ & $ \SI{3.2}{} $ & $ \SI{3.5}{} \pm \SI{1.1}{} $\\
	
    $\tilde{z} (\SI{}{\milli\meter})$   & $\SI{2.8}{}\pm \SI{0.9}{}$ & $ \SI{3.3}{} $&   $ \SI{3.4}{} \pm \SI{1.1}{} $ \\
    \bottomrule
    \end{tabular}
	\begin{tabular}{llll} 
    \toprule
	\multicolumn{4}{c}{f = \SI{100}{\hertz}}\\
    & S & TT-NP & TT-WP  \\
	& $\bar{\tilde{X}} \pm \sigma$ & RMSE& $\bar{\tilde{X}} \pm {\sigma}$   \\ 
	\midrule
	$\tilde{x} (\SI{}{\milli\meter})$ & $\SI{1.7}{}\pm \SI{0.6}{}$ & $ \SI{3.4}{} $& $ \SI{3.6}{} \pm \SI{1.2}{} $ \\ 
	
	$\tilde{y} (\SI{}{\milli\meter})$ &   $\SI{1.7}{}\pm \SI{0.5}{}$  & $ \SI{3.7}{} $&  $ \SI{3.9}{} \pm \SI{1.3}{} $\\
	
    $\tilde{z} (\SI{}{\milli\meter})$   & $\SI{2.6}{}\pm \SI{0.9}{}$& $ \SI{2.8}{} $ & $ \SI{3.0}{} \pm \SI{1.0}{} $ \\
\bottomrule
\end{tabular}
	\begin{tabular}{llll} 
	\\
	\toprule 
    \multicolumn{4}{c}{f = \SI{20}{\hertz}}\\
    & S & TT-NP & TT-WP  \\
	& $\bar{\tilde{X}} \pm \sigma$ & RMSE& $\bar{\tilde{X}} \pm {\sigma}$   \\ 
	\midrule
	$\tilde{x} (\SI{}{\milli\meter})$ & $\SI{1.8}{}\pm \SI{0.6}{}$ & $ \SI{3.7}{} $& $ \SI{3.9}{} \pm \SI{1.3}{} $ \\ 
	
	$\tilde{y} (\SI{}{\milli\meter})$  & $\SI{1.9}{}\pm 
	\SI{0.6}{}$ & $ \SI{4.2}{} $ &   $ \SI{4.6}{} \pm \SI{1.5}{} $ \\
	
    $\tilde{z} (\SI{}{\milli\meter})$  &  $\SI{2.8}{}\pm \SI{0.9}{}$ & $ \SI{3.3}{} $&   $ \SI{3.4}{} \pm \SI{1.1}{} $ \\
\bottomrule
\end{tabular}
\end{center}
\label{tab:4}
\end{table}
\addtolength{\belowcaptionskip}{0pt}
\section{Discussion} \label{sec:Discussion}
The experiments verify that the FIC retains passivity in a redundant manipulator without requiring the observation of the entire robot state. As a consequence, the FIC's stability is independent of the null-space controller, which is regarded as an external perturbation. The tracking accuracy data indicate that it is robust to model errors, which are equivalent to external disturbances and are compensated by the FIC's nonlinear stiffness. The experiments also show that the robot can initiate rigid contact with unknown bodies and dynamic systems without any assumption about the interaction.
In the attached video, the reader can also appreciate how the robot remains stable even in the case of a sudden loss of contact when applying the maximum achievable interaction force.
The robustness is further showcased in \autoref{fig:8} and \autoref{fig:9} where the robot initiates and brakes contact with 3 different rigid bodies without any need to account for the objects' presence or shape in the controller.
Furthermore, the video includes the following additional scenarios: FIC with damping in the convergence phase, perturbation beyond the boundaries without damping and rapid shaking without damping.
The passivity was verified in all cases. The time of the convergence are $\SI{1.24}{\second}$, $\SI{1.33}{\second}$ and $\SI{1.35}{\second}$, respectively.
Therefore, it can be said that the controller is intrinsically stable independently from environmental interactions both theoretically and experimentally.
The data also show that the trajectory tracking errors RMSE (\autoref{tab:2} and \autoref{tab:3}) are always contained within the task boundaries set by the user.
Thus, we validate that tracking performance is not affected by the removal of active damping due to the high online adjustable impedance gains that may be achieved with the proposed controller.

These results are possible because the stability of the proposed controller does not rely on the projection matrices used by conventional controllers to account for null-space and contact during stabilization \cite{franco,XinGetal, Moura}.
Mainly, \textit{Moura et al.} have recently studied and reported the effect of numerical instability related to projection matrices on the performance of both contact and null-space controllers, which is considered a major limitation of stable interaction with highly variable and unknown dynamics \cite{Moura}.

In summary, the major difference between tank-based controllers and the proposed method is that a nonlinear controller impedance as an energy storage function eliminates dependency on null-space projections and, consequently, singularities.
The FIC energy maps the robot energy, which is upper-bounded by the maximum displacement from the desired pose and is zero only when the robot is in the desired pose. Consequently, the FIC overcomes the limitation of the tank-based approach that is susceptible to miss-estimation of the exchanged energy \cite{passivationOfProjectionBased}. 

The experiments with low-feedback bandwidth demonstrate that the proposed method is immune to the miss-estimation of exchanged energy intrinsic to methods requiring integration to evaluate the state of the controller.  The tracking error is slightly increased by the reduction of the feedback bandwidth, and there is also an increase in the oscillatory behaviour while recovering from a perturbation. 
Thus, reduced feedback bandwidth degrades the system's dynamic response, but it does not alter the stability proprieties of the controller. 
These results are particularly important because they show how this controller may be operated directly via feedback from camera sensors, which typically require computationally complex sensor fusion and state estimators that are susceptible to drift \cite{sabatini2006quaternion,bloesch2013state,li2020visual}.

The sole limitation of the proposed method currently known to the authors is that when using a nonlinear impedance profile, there is a jump in force when switching from divergence to convergence. This behaviour is due to the different slopes of the stiffness energy in the two phases, and further research needs to be conducted to identify a solution to ensure energy and force conservation at the switching condition for nonlinear force profiles.

\section{CONCLUSION} \label{sec:Conclusions}
This manuscript introduces a framework for impedance controllers which relies on a fractal task-space attractor to guarantee the stability of interaction with unknown environments.  
The FIC enables online impedance adaptation without introducing additional stability conditions, which is not possible with traditional controllers \cite{franco,AVersatileBiomimeticCtrl,nonSmooth}.
The results show that the system can achieve a good level of accuracy in trajectory tracking, and it can exert significant forces on the environment without compromising stability. To the best of the authors' knowledge, these results are the first example of accurate trajectory tracking that relies solely upon the nonlinear stiffness of the controller.
Future works will focus on identifying different impedance profiles and attractor characteristics that will enable the application of this framework to fields such as medical and industrial robots.

\section*{Declarations}
\subsection*{Conflict of Interest}
The authors declare that they have no conflict of interest.
\subsection*{Funding}
This work has been supported by the following grants: EPSRC UK RAI Hub ORCA (EP/R026173/1) and NCNR (EP/R02572X/1), CogIMon project in the EU Horizon 2020 (ICT-23-2014), THING project in the EU Horizon 2020 (ICT-2017-1), and by grant EP/L016834/1 for the University of Edinburgh RAS CDT from EPSRC.
\subsection*{Supporting Data}
The data that support the findings of this study are available from the corresponding authors upon reasonable request.

\bibliography{sn-bibliography}


\begin{thebibliography}{45}
\ifx \bisbn   \undefined \def \bisbn  #1{ISBN #1}\fi
\ifx \binits  \undefined \def \binits#1{#1}\fi
\ifx \bauthor  \undefined \def \bauthor#1{#1}\fi
\ifx \batitle  \undefined \def \batitle#1{#1}\fi
\ifx \bjtitle  \undefined \def \bjtitle#1{#1}\fi
\ifx \bvolume  \undefined \def \bvolume#1{\textbf{#1}}\fi
\ifx \byear  \undefined \def \byear#1{#1}\fi
\ifx \bissue  \undefined \def \bissue#1{#1}\fi
\ifx \bfpage  \undefined \def \bfpage#1{#1}\fi
\ifx \blpage  \undefined \def \blpage #1{#1}\fi
\ifx \burl  \undefined \def \burl#1{\textsf{#1}}\fi
\ifx \doiurl  \undefined \def \doiurl#1{\url{https://doi.org/#1}}\fi
\ifx \betal  \undefined \def \betal{\textit{et al.}}\fi
\ifx \binstitute  \undefined \def \binstitute#1{#1}\fi
\ifx \binstitutionaled  \undefined \def \binstitutionaled#1{#1}\fi
\ifx \bctitle  \undefined \def \bctitle#1{#1}\fi
\ifx \beditor  \undefined \def \beditor#1{#1}\fi
\ifx \bpublisher  \undefined \def \bpublisher#1{#1}\fi
\ifx \bbtitle  \undefined \def \bbtitle#1{#1}\fi
\ifx \bedition  \undefined \def \bedition#1{#1}\fi
\ifx \bseriesno  \undefined \def \bseriesno#1{#1}\fi
\ifx \blocation  \undefined \def \blocation#1{#1}\fi
\ifx \bsertitle  \undefined \def \bsertitle#1{#1}\fi
\ifx \bsnm \undefined \def \bsnm#1{#1}\fi
\ifx \bsuffix \undefined \def \bsuffix#1{#1}\fi
\ifx \bparticle \undefined \def \bparticle#1{#1}\fi
\ifx \barticle \undefined \def \barticle#1{#1}\fi
\bibcommenthead
\ifx \bconfdate \undefined \def \bconfdate #1{#1}\fi
\ifx \botherref \undefined \def \botherref #1{#1}\fi
\ifx \url \undefined \def \url#1{\textsf{#1}}\fi
\ifx \bchapter \undefined \def \bchapter#1{#1}\fi
\ifx \bbook \undefined \def \bbook#1{#1}\fi
\ifx \bcomment \undefined \def \bcomment#1{#1}\fi
\ifx \oauthor \undefined \def \oauthor#1{#1}\fi
\ifx \citeauthoryear \undefined \def \citeauthoryear#1{#1}\fi
\ifx \endbibitem  \undefined \def \endbibitem {}\fi
\ifx \bconflocation  \undefined \def \bconflocation#1{#1}\fi
\ifx \arxivurl  \undefined \def \arxivurl#1{\textsf{#1}}\fi
\csname PreBibitemsHook\endcsname

\bibitem{impCtrl}
\begin{botherref}
\oauthor{\bsnm{Hogan}, \binits{N.}}:
Impedance control: An approach to manipulation.
Journal of dynamic systems, measurement, and control
\textbf{107}(17)
(1985)
\end{botherref}
\endbibitem

\bibitem{Hogan2014}
\begin{barticle}
\bauthor{\bsnm{Hogan}, \binits{N.}}:
\batitle{A general actuator model based on nonlinear equivalent networks}.
\bjtitle{IEEE/ASME Transactions on Mechatronics}
\bvolume{19}(\bissue{6}),
\bfpage{1929}--\blpage{1939}
(\byear{2013})
\end{barticle}
\endbibitem

\bibitem{capelli2022passivity}
\begin{botherref}
\oauthor{\bsnm{Capelli}, \binits{B.}},
\oauthor{\bsnm{Secchi}, \binits{C.}},
\oauthor{\bsnm{Sabattini}, \binits{L.}}:
Passivity and control barrier functions: Optimizing the use of energy.
IEEE Robotics and Automation Letters
(2022)
\end{botherref}
\endbibitem

\bibitem{lachner2021energy}
\begin{barticle}
\bauthor{\bsnm{Lachner}, \binits{J.}},
\bauthor{\bsnm{Allmendinger}, \binits{F.}},
\bauthor{\bsnm{Hobert}, \binits{E.}},
\bauthor{\bsnm{Hogan}, \binits{N.}},
\bauthor{\bsnm{Stramigioli}, \binits{S.}}:
\batitle{Energy budgets for coordinate invariant robot control in physical
  human--robot interaction}.
\bjtitle{The International Journal of Robotics Research}
\bvolume{40}(\bissue{8-9}),
\bfpage{968}--\blpage{985}
(\byear{2021})
\end{barticle}
\endbibitem

\bibitem{lachner2022shaping}
\begin{botherref}
\oauthor{\bsnm{Lachner}, \binits{J.}},
\oauthor{\bsnm{Allmendinger}, \binits{F.}},
\oauthor{\bsnm{Stramigioli}, \binits{S.}},
\oauthor{\bsnm{Hogan}, \binits{N.}}:
Shaping impedances to comply with constrained task dynamics.
IEEE Transactions on Robotics
(2022)
\end{botherref}
\endbibitem

\bibitem{ramuzat2022passive}
\begin{botherref}
\oauthor{\bsnm{Ramuzat}, \binits{N.}},
\oauthor{\bsnm{Boria}, \binits{S.}},
\oauthor{\bsnm{Stasse}, \binits{O.}}:
Passive inverse dynamics control using a global energy tank for
  torque-controlled humanoid robots in multi-contact.
IEEE Robotics and Automation Letters
(2022)
\end{botherref}
\endbibitem

\bibitem{smith2002synthesis}
\begin{barticle}
\bauthor{\bsnm{Smith}, \binits{M.C.}}:
\batitle{Synthesis of mechanical networks: the inerter}.
\bjtitle{IEEE Transactions on automatic control}
\bvolume{47}(\bissue{10}),
\bfpage{1648}--\blpage{1662}
(\byear{2002})
\end{barticle}
\endbibitem

\bibitem{forceImpTrajLearning}
\begin{barticle}
\bauthor{\bsnm{Li}, \binits{Y.}},
\bauthor{\bsnm{Ganesh}, \binits{G.}},
\bauthor{\bsnm{Jarrass{\'e}}, \binits{N.}},
\bauthor{\bsnm{Haddadin}, \binits{S.}},
\bauthor{\bsnm{Albu-Schaeffer}, \binits{A.}},
\bauthor{\bsnm{Burdet}, \binits{E.}}:
\batitle{Force, impedance, and trajectory learning for contact tooling and
  haptic identification}.
\bjtitle{IEEE Transactions on Robotics}
\bvolume{34}(\bissue{5}),
\bfpage{1170}--\blpage{1182}
(\byear{2018})
\end{barticle}
\endbibitem

\bibitem{varImpCtrlBasedOnHumanStiffEstimation}
\begin{bchapter}
\bauthor{\bsnm{Tsumugiwa}, \binits{T.}},
\bauthor{\bsnm{Yokogawa}, \binits{R.}},
\bauthor{\bsnm{Hara}, \binits{K.}}:
\bctitle{Variable impedance control based on estimation of human arm stiffness
  for human-robot cooperative calligraphic task}.
In: \bbtitle{Proceedings 2002 IEEE International Conference on Robotics and
  Automation (Cat. No. 02CH37292)},
vol. \bseriesno{1},
pp. \bfpage{644}--\blpage{650}
(\byear{2002}).
\bcomment{IEEE}
\end{bchapter}
\endbibitem

\bibitem{lin2018projected}
\begin{bchapter}
\bauthor{\bsnm{Lin}, \binits{H.-C.}},
\bauthor{\bsnm{Smith}, \binits{J.}},
\bauthor{\bsnm{Babarahmati}, \binits{K.K.}},
\bauthor{\bsnm{Dehio}, \binits{N.}},
\bauthor{\bsnm{Mistry}, \binits{M.}}:
\bctitle{A projected inverse dynamics approach for multi-arm cartesian
  impedance control}.
In: \bbtitle{IEEE International Conference on Robotics and Automation (ICRA)},
pp. \bfpage{1}--\blpage{5}
(\byear{2018}).
\bcomment{IEEE}
\end{bchapter}
\endbibitem

\bibitem{HumanLikeAdaptOfForceAndImp}
\begin{barticle}
\bauthor{\bsnm{Yang}, \binits{C.}},
\bauthor{\bsnm{Ganesh}, \binits{G.}},
\bauthor{\bsnm{Haddadin}, \binits{S.}},
\bauthor{\bsnm{Parusel}, \binits{S.}},
\bauthor{\bsnm{Albu-Schaeffer}, \binits{A.}},
\bauthor{\bsnm{Burdet}, \binits{E.}}:
\batitle{Human-like adaptation of force and impedance in stable and unstable
  interactions}.
\bjtitle{IEEE transactions on robotics}
\bvolume{27}(\bissue{5}),
\bfpage{918}--\blpage{930}
(\byear{2011})
\end{barticle}
\endbibitem

\bibitem{SU2020}
\begin{bchapter}
\bauthor{\bsnm{{Su}}, \binits{H.}},
\bauthor{\bsnm{{Schmirander}}, \binits{Y.}},
\bauthor{\bsnm{{Li}}, \binits{Z.}},
\bauthor{\bsnm{{Zhou}}, \binits{X.}},
\bauthor{\bsnm{{Ferrigno}}, \binits{G.}},
\bauthor{\bsnm{{De Momi}}, \binits{E.}}:
\bctitle{Bilateral teleoperation control of a redundant manipulator with an rcm
  kinematic constraint}.
In: \bbtitle{2020 IEEE International Conference on Robotics and Automation
  (ICRA)},
pp. \bfpage{4477}--\blpage{4482}
(\byear{2020}).
\doiurl{10.1109/ICRA40945.2020.9197267}
\end{bchapter}
\endbibitem

\bibitem{Spyrakos-Papastavridis2020}
\begin{barticle}
\bauthor{\bsnm{{Spyrakos-Papastavridis}}, \binits{E.}},
\bauthor{\bsnm{{Childs}}, \binits{P.R.N.}},
\bauthor{\bsnm{{Dai}}, \binits{J.S.}}:
\batitle{Passivity preservation for variable impedance control of compliant
  robots}.
\bjtitle{IEEE/ASME Transactions on Mechatronics}
\bvolume{25}(\bissue{5}),
\bfpage{2342}--\blpage{2353}
(\byear{2020}).
\doiurl{10.1109/TMECH.2019.2961478}
\end{barticle}
\endbibitem

\bibitem{varImpCtrlReinforcementLearApproach}
\begin{botherref}
\oauthor{\bsnm{Buchli}, \binits{J.}},
\oauthor{\bsnm{Theodorou}, \binits{E.}},
\oauthor{\bsnm{Stulp}, \binits{F.}},
\oauthor{\bsnm{Schaal}, \binits{S.}}:
Variable impedance control a reinforcement learning approach.
Robotics: Science and Systems VI,
153--160
(2011)
\end{botherref}
\endbibitem

\bibitem{TankBasedApproachImpCtrlVarStiff}
\begin{bchapter}
\bauthor{\bsnm{Ferraguti}, \binits{F.}},
\bauthor{\bsnm{Secchi}, \binits{C.}},
\bauthor{\bsnm{Fantuzzi}, \binits{C.}}:
\bctitle{A tank-based approach to impedance control with variable stiffness}.
In: \bbtitle{2013 IEEE International Conference on Robotics and Automation},
pp. \bfpage{4948}--\blpage{4953}
(\byear{2013}).
\bcomment{IEEE}
\end{bchapter}
\endbibitem

\bibitem{impCtrlWithVarStiffGains}
\begin{bchapter}
\bauthor{\bsnm{Park}, \binits{J.H.}},
\bauthor{\bsnm{Cho}, \binits{H.C.}}:
\bctitle{Impedance control with varying stiffness for parallel-link
  manipulators}.
In: \bbtitle{Proceedings of the 1998 American Control Conference. ACC (IEEE
  Cat. No. 98CH36207)},
vol. \bseriesno{1},
pp. \bfpage{478}--\blpage{482}
(\byear{1998}).
\bcomment{IEEE}
\end{bchapter}
\endbibitem

\bibitem{AVersatileBiomimeticCtrl}
\begin{bchapter}
\bauthor{\bsnm{Ganesh}, \binits{G.}},
\bauthor{\bsnm{Jarrass{\'e}}, \binits{N.}},
\bauthor{\bsnm{Haddadin}, \binits{S.}},
\bauthor{\bsnm{Albu-Schaeffer}, \binits{A.}},
\bauthor{\bsnm{Burdet}, \binits{E.}}:
\bctitle{A versatile biomimetic controller for contact tooling and haptic
  exploration}.
In: \bbtitle{2012 IEEE International Conference on Robotics and Automation},
pp. \bfpage{3329}--\blpage{3334}
(\byear{2012}).
\bcomment{IEEE}
\end{bchapter}
\endbibitem

\bibitem{zhao2020asymmetrical}
\begin{barticle}
\bauthor{\bsnm{Zhao}, \binits{X.}},
\bauthor{\bsnm{Tao}, \binits{B.}},
\bauthor{\bsnm{Qian}, \binits{L.}},
\bauthor{\bsnm{Yang}, \binits{Y.}},
\bauthor{\bsnm{Ding}, \binits{H.}}:
\batitle{Asymmetrical nonlinear impedance control for dual robotic machining of
  thin-walled workpieces}.
\bjtitle{Robotics and Computer-Integrated Manufacturing}
\bvolume{63},
\bfpage{101889}
(\byear{2020})
\end{barticle}
\endbibitem

\bibitem{dong2019adaptive}
\begin{barticle}
\bauthor{\bsnm{Dong}, \binits{G.}},
\bauthor{\bsnm{Huang}, \binits{P.}},
\bauthor{\bsnm{Ma}, \binits{Z.}}:
\batitle{Adaptive stiffness and damping impedance control for environmental
  interactive systems with unknown uncertainty and disturbance}.
\bjtitle{IEEE Access}
\bvolume{7},
\bfpage{172433}--\blpage{172442}
(\byear{2019})
\end{barticle}
\endbibitem

\bibitem{ForceTrackImpCtrl}
\begin{barticle}
\bauthor{\bsnm{Lee}, \binits{K.}},
\bauthor{\bsnm{Buss}, \binits{M.}}:
\batitle{Force tracking impedance control with variable target stiffness}.
\bjtitle{IFAC Proceedings Volumes}
\bvolume{41}(\bissue{2}),
\bfpage{6751}--\blpage{6756}
(\byear{2008})
\end{barticle}
\endbibitem

\bibitem{S1}
\begin{barticle}
\bauthor{\bsnm{Erden}, \binits{M.S.}},
\bauthor{\bsnm{Billard}, \binits{A.}}:
\batitle{Robotic assistance by impedance compensation for hand movements while
  manual welding}.
\bjtitle{IEEE transactions on cybernetics}
\bvolume{46}(\bissue{11}),
\bfpage{2459}--\blpage{2472}
(\byear{2015})
\end{barticle}
\endbibitem

\bibitem{S2}
\begin{barticle}
\bauthor{\bsnm{Erden}, \binits{M.S.}},
\bauthor{\bsnm{Billard}, \binits{A.}}:
\batitle{End-point impedance measurements across dominant and nondominant hands
  and robotic assistance with directional damping}.
\bjtitle{IEEE transactions on cybernetics}
\bvolume{45}(\bissue{6}),
\bfpage{1146}--\blpage{1157}
(\byear{2014})
\end{barticle}
\endbibitem

\bibitem{S5}
\begin{bchapter}
\bauthor{\bsnm{Tugal}, \binits{H.}},
\bauthor{\bsnm{Gautier}, \binits{B.}},
\bauthor{\bsnm{Kircicek}, \binits{M.}},
\bauthor{\bsnm{Erden}, \binits{M.S.}}:
\bctitle{Hand-impedance measurement during laparoscopic training coupled with
  robotic manipulators}.
In: \bbtitle{2018 IEEE/RSJ International Conference on Intelligent Robots and
  Systems (IROS)},
pp. \bfpage{4404}--\blpage{4410}
(\byear{2018}).
\bcomment{IEEE}
\end{bchapter}
\endbibitem

\bibitem{bilateralTeleManWithTimeDelays}
\begin{barticle}
\bauthor{\bsnm{Franken}, \binits{M.}},
\bauthor{\bsnm{Stramigioli}, \binits{S.}},
\bauthor{\bsnm{Misra}, \binits{S.}},
\bauthor{\bsnm{Secchi}, \binits{C.}},
\bauthor{\bsnm{Macchelli}, \binits{A.}}:
\batitle{Bilateral telemanipulation with time delays: A two-layer approach
  combining passivity and transparency}.
\bjtitle{IEEE transactions on robotics}
\bvolume{27}(\bissue{4}),
\bfpage{741}--\blpage{756}
(\byear{2011})
\end{barticle}
\endbibitem

\bibitem{portBasedAsymptoticCurveTracking}
\begin{barticle}
\bauthor{\bsnm{Duindam}, \binits{V.}},
\bauthor{\bsnm{Stramigioli}, \binits{S.}}:
\batitle{Port-based asymptotic curve tracking for mechanical systems}.
\bjtitle{European Journal of Control}
\bvolume{10}(\bissue{5}),
\bfpage{411}--\blpage{420}
(\byear{2004})
\end{barticle}
\endbibitem

\bibitem{energyTankBasedWrenchImpCtrl}
\begin{bchapter}
\bauthor{\bsnm{{Rashad}}, \binits{R.}},
\bauthor{\bsnm{{Engelen}}, \binits{J.B.C.}},
\bauthor{\bsnm{{Stramigioli}}, \binits{S.}}:
\bctitle{Energy tank-based wrench/impedance control of a fully-actuated
  hexarotor: A geometric port-hamiltonian approach}.
In: \bbtitle{2019 International Conference on Robotics and Automation (ICRA)},
pp. \bfpage{6418}--\blpage{6424}
(\byear{2019}).
\doiurl{10.1109/ICRA.2019.8793939}
\end{bchapter}
\endbibitem

\bibitem{PowerFlowRegulationAdaptLearnVirtEnergyTank}
\begin{barticle}
\bauthor{\bsnm{{Shahriari}}, \binits{E.}},
\bauthor{\bsnm{{Johannsmeier}}, \binits{L.}},
\bauthor{\bsnm{{Jensen}}, \binits{E.}},
\bauthor{\bsnm{{Haddadin}}, \binits{S.}}:
\batitle{Power flow regulation, adaptation, and learning for intrinsically
  robust virtual energy tanks}.
\bjtitle{IEEE Robotics and Automation Letters}
\bvolume{5}(\bissue{1}),
\bfpage{211}--\blpage{218}
(\byear{2020}).
\doiurl{10.1109/LRA.2019.2953662}
\end{barticle}
\endbibitem

\bibitem{positionDriftCompensation}
\begin{bchapter}
\bauthor{\bsnm{Secchi}, \binits{C.}},
\bauthor{\bsnm{Stramigioli}, \binits{S.}},
\bauthor{\bsnm{Fantuzzi}, \binits{C.}}:
\bctitle{Position drift compensation in port-hamiltonian based
  telemanipulation}.
In: \bbtitle{2006 IEEE/RSJ International Conference on Intelligent Robots and
  Systems},
pp. \bfpage{4211}--\blpage{4216}
(\byear{2006}).
\bcomment{IEEE}
\end{bchapter}
\endbibitem

\bibitem{khoramshahi2020dynamical}
\begin{barticle}
\bauthor{\bsnm{Khoramshahi}, \binits{M.}},
\bauthor{\bsnm{Billard}, \binits{A.}}:
\batitle{A dynamical system approach for detection and reaction to human
  guidance in physical human--robot interaction}.
\bjtitle{Autonomous Robots}
\bvolume{44}(\bissue{8}),
\bfpage{1411}--\blpage{1429}
(\byear{2020})
\end{barticle}
\endbibitem

\bibitem{unifiedPassivityBased}
\begin{bchapter}
\bauthor{\bsnm{Schindlbeck}, \binits{C.}},
\bauthor{\bsnm{Haddadin}, \binits{S.}}:
\bctitle{Unified passivity-based cartesian force/impedance control for rigid
  and flexible joint robots via task-energy tanks}.
In: \bbtitle{2015 IEEE International Conference on Robotics and Automation
  (ICRA)},
pp. \bfpage{440}--\blpage{447}
(\byear{2015}).
\bcomment{IEEE}
\end{bchapter}
\endbibitem

\bibitem{passivationOfProjectionBased}
\begin{barticle}
\bauthor{\bsnm{Dietrich}, \binits{A.}},
\bauthor{\bsnm{Ott}, \binits{C.}},
\bauthor{\bsnm{Stramigioli}, \binits{S.}}:
\batitle{Passivation of projection-based null space compliance control via
  energy tanks}.
\bjtitle{IEEE Robotics and automation letters}
\bvolume{1}(\bissue{1}),
\bfpage{184}--\blpage{191}
(\byear{2015})
\end{barticle}
\endbibitem

\bibitem{stabilityConsiderationsVariImpCtrl}
\begin{barticle}
\bauthor{\bsnm{Kronander}, \binits{K.}},
\bauthor{\bsnm{Billard}, \binits{A.}}:
\batitle{Stability considerations for variable impedance control}.
\bjtitle{IEEE Transactions on Robotics}
\bvolume{32}(\bissue{5}),
\bfpage{1298}--\blpage{1305}
(\byear{2016})
\end{barticle}
\endbibitem

\bibitem{combiningEnergyAndPowerBasedSafety}
\begin{bchapter}
\bauthor{\bsnm{Tadele}, \binits{T.S.}},
\bauthor{\bparticle{de} \bsnm{Vries}, \binits{T.J.}},
\bauthor{\bsnm{Stramigioli}, \binits{S.}}:
\bctitle{Combining energy and power based safety metrics in controller design
  for domestic robots}.
In: \bbtitle{2014 IEEE International Conference on Robotics and Automation
  (ICRA)},
pp. \bfpage{1209}--\blpage{1214}
(\byear{2014}).
\bcomment{IEEE}
\end{bchapter}
\endbibitem

\bibitem{safetyAndEnergyAwareImpCtrl}
\begin{barticle}
\bauthor{\bsnm{Raiola}, \binits{G.}},
\bauthor{\bsnm{Cardenas}, \binits{C.A.}},
\bauthor{\bsnm{Tadele}, \binits{T.S.}},
\bauthor{\bsnm{De~Vries}, \binits{T.}},
\bauthor{\bsnm{Stramigioli}, \binits{S.}}:
\batitle{Development of a safety-and energy-aware impedance controller for
  collaborative robots}.
\bjtitle{IEEE Robotics and automation letters}
\bvolume{3}(\bissue{2}),
\bfpage{1237}--\blpage{1244}
(\byear{2018})
\end{barticle}
\endbibitem

\bibitem{stramigioli2005}
\begin{barticle}
\bauthor{\bsnm{Stramigioli}, \binits{S.}},
\bauthor{\bsnm{Secchi}, \binits{C.}},
\bauthor{\bparticle{van~der} \bsnm{Schaft}, \binits{A.J.}},
\bauthor{\bsnm{Fantuzzi}, \binits{C.}}:
\batitle{Sampled data systems passivity and discrete port-hamiltonian systems}.
\bjtitle{IEEE Transactions on Robotics}
\bvolume{21}(\bissue{4}),
\bfpage{574}--\blpage{587}
(\byear{2005})
\end{barticle}
\endbibitem

\bibitem{Sciavicco}
\begin{bbook}
\bauthor{\bsnm{Sciavicco}, \binits{L.}},
\bauthor{\bsnm{Siciliano}, \binits{B.}}:
\bbtitle{Modelling and Control of Robot Manipulators}.
\bpublisher{Springer},
\blocation{London, UK}
(\byear{2012})
\end{bbook}
\endbibitem

\bibitem{strogatz2018nonlinear}
\begin{bbook}
\bauthor{\bsnm{Strogatz}, \binits{S.H.}}:
\bbtitle{Nonlinear Dynamics and Chaos with Student Solutions Manual: With
  Applications to Physics, Biology, Chemistry, and Engineering}.
\bpublisher{CRC press},
\blocation{Boca Raton, FL, USA}
(\byear{2018})
\end{bbook}
\endbibitem

\bibitem{SwitchingArchitecture}
\begin{bchapter}
\bauthor{\bsnm{Swift}, \binits{T.A.}},
\bauthor{\bsnm{Strausser}, \binits{K.A.}},
\bauthor{\bsnm{Zoss}, \binits{A.B.}},
\bauthor{\bsnm{Kazerooni}, \binits{H.}}:
\bctitle{Control and experimental results for post stroke gait rehabilitation
  with a prototype mobile medical exoskeleton}.
In: \bbtitle{ASME 2010 Dynamic Systems and Control Conference},
pp. \bfpage{405}--\blpage{411}
(\byear{2010}).
\bcomment{American Society of Mechanical Engineers}
\end{bchapter}
\endbibitem

\bibitem{franco}
\begin{bchapter}
\bauthor{\bsnm{Angelini}, \binits{F.}},
\bauthor{\bsnm{Xin}, \binits{G.}},
\bauthor{\bsnm{Wolfslag}, \binits{W.J.}},
\bauthor{\bsnm{Tiseo}, \binits{C.}},
\bauthor{\bsnm{Mistry}, \binits{M.}},
\bauthor{\bsnm{Garabini}, \binits{M.}},
\bauthor{\bsnm{Bicchi}, \binits{A.}},
\bauthor{\bsnm{Vijayakumar}, \binits{S.}}:
\bctitle{Online optimal impedance planning for legged robots}.
In: \bbtitle{IEEE International Conference on Intelligent Robots and Systems},
pp. \bfpage{6028}--\blpage{6035}
(\byear{2019}).
\doiurl{10.1109/IROS40897.2019.8967696}
\end{bchapter}
\endbibitem

\bibitem{XinGetal}
\begin{barticle}
\bauthor{\bsnm{Xin}, \binits{G.}},
\bauthor{\bsnm{Wolfslag}, \binits{W.}},
\bauthor{\bsnm{Lin}, \binits{H.-C.}},
\bauthor{\bsnm{Tiseo}, \binits{C.}},
\bauthor{\bsnm{Mistry}, \binits{M.}}:
\batitle{An optimization-based locomotion controller for quadruped robots
  leveraging cartesian impedance control}.
\bjtitle{Frontiers in Robotics and AI}
\bvolume{7},
\bfpage{48}
(\byear{2020})
\end{barticle}
\endbibitem

\bibitem{Moura}
\begin{bchapter}
\bauthor{\bsnm{Moura}, \binits{J.}},
\bauthor{\bsnm{Ivan}, \binits{V.}},
\bauthor{\bsnm{Erden}, \binits{M.S.}},
\bauthor{\bsnm{Vijayakumar}, \binits{S.}}:
\bctitle{Equivalence of the projected forward dynamics and the dynamically
  consistent inverse solution}.
In: \bbtitle{2019 Robotics: Science and Systems},
p. \bfpage{00}
(\byear{2019})
\end{bchapter}
\endbibitem

\bibitem{sabatini2006quaternion}
\begin{barticle}
\bauthor{\bsnm{Sabatini}, \binits{A.M.}}:
\batitle{Quaternion-based extended kalman filter for determining orientation by
  inertial and magnetic sensing}.
\bjtitle{IEEE transactions on Biomedical Engineering}
\bvolume{53}(\bissue{7}),
\bfpage{1346}--\blpage{1356}
(\byear{2006})
\end{barticle}
\endbibitem

\bibitem{bloesch2013state}
\begin{barticle}
\bauthor{\bsnm{Bloesch}, \binits{M.}},
\bauthor{\bsnm{Hutter}, \binits{M.}},
\bauthor{\bsnm{Hoepflinger}, \binits{M.A.}},
\bauthor{\bsnm{Leutenegger}, \binits{S.}},
\bauthor{\bsnm{Gehring}, \binits{C.}},
\bauthor{\bsnm{Remy}, \binits{C.D.}},
\bauthor{\bsnm{Siegwart}, \binits{R.}}:
\batitle{State estimation for legged robots-consistent fusion of leg kinematics
  and imu}.
\bjtitle{Robotics}
\bvolume{17},
\bfpage{17}--\blpage{24}
(\byear{2013})
\end{barticle}
\endbibitem

\bibitem{li2020visual}
\begin{botherref}
\oauthor{\bsnm{Li}, \binits{S.}},
\oauthor{\bparticle{van~der} \bsnm{Horst}, \binits{E.}},
\oauthor{\bsnm{Duernay}, \binits{P.}},
\oauthor{\bsnm{De~Wagter}, \binits{C.}},
\oauthor{\bparticle{de} \bsnm{Croon}, \binits{G.C.}}:
Visual model-predictive localization for computationally efficient autonomous
  racing of a 72-g drone.
Journal of Field Robotics
(2020)
\end{botherref}
\endbibitem

\bibitem{nonSmooth}
\begin{barticle}
\bauthor{\bsnm{Shevitz}, \binits{D.}},
\bauthor{\bsnm{Paden}, \binits{B.}}:
\batitle{Lyapunov stability theory of nonsmooth systems}.
\bjtitle{IEEE Transactions on automatic control}
\bvolume{39}(\bissue{9}),
\bfpage{1910}--\blpage{1914}
(\byear{1994})
\end{barticle}
\endbibitem

\end{thebibliography}

\end{document}